%% file: elsarticle-template-harv.tex
\documentclass[final,3p,times,twocolumn,authoryear]{elsarticle}

\usepackage{amsmath,amssymb}
\usepackage{algorithmic}
\usepackage{algorithm}

\newcommand{\argmin}{\mathop{\rm argmin}\limits}
\usepackage{multirow}
\usepackage{multicol}
\usepackage{siunitx}
\usepackage{bbm}
\usepackage{comment}
\usepackage{graphicx}
\usepackage{subcaption}
\usepackage{siunitx}

\journal{Neural Networks}

\begin{document}

\begin{frontmatter}



\title{Goal-Conditioned Terminal Value Estimation for Real-time and Multi-task Model Predictive Control}


\affiliation[KU]{organization={Graduate School of Informatics, Kyoto University},
            addressline={Yoshida-honmachi 36-1}, 
            city={Sakyo-ku, Kyoto},
            postcode={606-8501}, 
            country={Japan}}

\affiliation[ATR]{organization={Computational Neuroscience Labs, Advanced Telecomunication Research International (ATR)},
            addressline={2-2-2 Hikaridai, Seika-cho}, 
            city={Soraku-gun, Kyoto},
            postcode={619-0288}, 
            country={Japan}}

\author[KU]{Mitsuki Morita}
\author[ATR]{Satoshi Yamamori}
\author[KU]{Satoshi Yagi}
\author[ATR]{Norikazu Sugimoto}
\author[KU,ATR]{Jun Morimoto}

\begin{abstract}
While MPC enables nonlinear feedback control by solving an optimal control problem at each timestep, the computational burden tends to be significantly large, making it difficult to optimize a policy within the control period. To address this issue, one possible approach is to utilize terminal value learning to reduce computational costs. However, the learned value cannot be used for other tasks in situations where the task dynamically changes in the original MPC setup. In this study, we develop an MPC framework with goal-conditioned terminal value learning to achieve multi-task policy optimization while reducing computational time. Furthermore, by using a hierarchical control structure that allows the upper-level trajectory planner to output appropriate goal-conditioned trajectories, we demonstrate that a robot model is able to generate diverse motions. We evaluate the proposed method on a bipedal inverted pendulum robot model and confirm that combining goal-conditioned terminal value learning with an upper-level trajectory planner enables real-time control; thus, the robot successfully tracks a target trajectory on sloped terrain.
\end{abstract}




\end{frontmatter}


\section{Introduction} \label{sec:intro}
\input{section/1introduction.tex}

\section{Related works} \label{sec:related_work}
\input{section/2related_works.tex}
\section{Preliminaries} \label{sec:preliminaries}
\input{section/3preliminaries}

\section{Method} \label{sec:method}
\input{section/4method}
\section{Experiments} \label{sec:experiments}
\input{section/5experiments}
\section{Results} \label{sec:results}
\input{section/6results}
\section{Conclusion} \label{sec:conclusion}
\input{section/7conclusion}







\section*{Acknowledgement}
This work was supported by JST-Mirai Program, Grant Number: JPMJMI21B1, JST Moonshot R\&D, Grant Number: JPMJMS223B-3,  JSPS KAKENHI Grant Number: 22H04998 and 22H03669, and the project JPNP20006, commissioned by NEDO, JAPAN.

\bibliographystyle{elsarticle-harv} 
\bibliography{reference} 

\clearpage
\appendix
\section{Hyperparameter} \label{sec:appendix_hyperparameter}
\input{section/8appendix_hyperparameter}

\end{document}

%% file: section/1introduction.tex
\begin{figure*}[h]
\centering
\includegraphics[width=\linewidth]{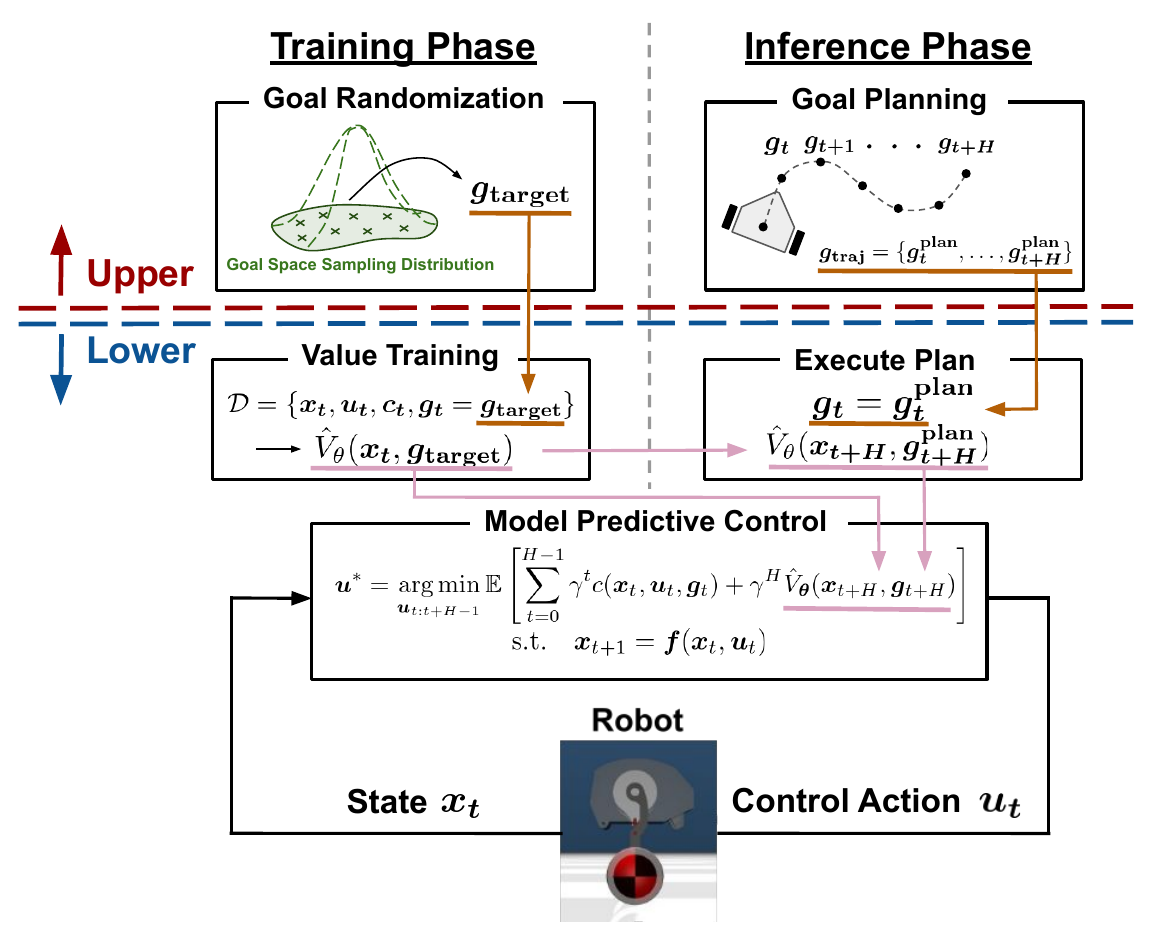}
\caption{
Schematic diagram of the proposed method. In the training phase, target goals $\boldsymbol{g}_\mathrm{target}$ are sampled from a predefined distribution, and a goal-conditioned value function $\hat{V}_{\theta}(\boldsymbol{x}_t, \boldsymbol{g}_{\mathrm{target}})$ is learned using MPC solutions. In the inference phase, the goal trajectory $\boldsymbol{g}^{\mathrm{plan}}_{t+H}$ is generated through planning for the MPC horizon. By inputting this trajectory into the MPC, the system can adapt to environmental changes and smoothly adjust its behavior. Here, $\boldsymbol{x}_t$, $\boldsymbol{u}_t$, $\boldsymbol{c}_t$, and $\boldsymbol{f}$ represent the state, control input, cost function, and system dynamics, respectively.}
\label{fig:research_overview}
\end{figure*}

In recent years, robot control and autonomous driving have been receiving a lot of attention. For such control targets, Model Predictive Control (MPC) \citep{ohtsuka_mpc, tassa2014control} is considered a promising approach. MPC provides a nonlinear feedback controller that computes the optimal control commands by solving an optimal control problem at each timestep, predicting the finite future states of the robot.  
Solving the optimal control problem is computationally demanding when the control system involves high-dimensional, multi-degree-of-freedom systems or operates in complex environments. This may hinder the completion of computations within the control cycle and prevent the achievement of control objectives.

Therefore, efforts have been made to approximate such complex robot systems with simpler models to solve the optimal control problem within a limited time and achieve control objectives \citep{orin, sleiman2021unified, farshidian2017efficient, SRBDMPC, humanoid_centoroidal}. However, assumptions such as the limbs being significantly lighter than the torso \citep{orin, SRBDMPC, humanoid_centoroidal}, or the instantaneous controllability of joint angular velocities \citep{sleiman2021unified, farshidian2017efficient}, are often made. Consequently, large modeling errors can lead to degraded performance or generate infeasible solutions.

Another approach to reducing MPC's computational costs involves shortening the prediction horizon, which refers to the length of future robot state predictions, by using terminal value learning \citep{polo, mpq, mpqlam}. While a longer prediction horizon improves control performance, it also increases computational costs. Terminal value learning allows for a shorter prediction horizon without significant performance loss. However, since the learned terminal values are task-specific, this approach limits the flexibility to dynamically change control targets, thus restricting the diversity of motions generated by MPC.

This study proposes a flexible control method that can change the control target without greatly simplifying the robot model while keeping the computational cost of MPC low. By adopting the idea of goal-conditioned reinforcement learning, our proposed method learns the terminal value with goal-related variables as inputs. Furthermore, we propose a hierarchical control architecture in which the MPC using the learned terminal values serves as the lower-layer controller and the trajectory generator that produces the target condition sequences acts as the upper-layer controller (see also Fig. \ref{fig:research_overview}). The proposed hierarchical approach allows for dynamic switching of objectives in response to the surrounding context, thus generating flexible and diverse robot behaviors. To evaluate our proposed method on real-time dynamic control tasks, we applied the method to a simulated model of a bipedal inverted pendulum robot. We successfully achieved real-time control for tasks such as following a lemniscate trajectory on sloped terrain.

This study includes the following contributions:
\begin{itemize}
      \item We propose a method to tackle the trade-off between flexibility and computational efficiency in model predictive control and value function learning for real-time and multi-task control problems.
      \item The proposed two-stage learning framework with domain randomization for training the hierarchical control strategy achieves a more flexible but faster derivation of the optimal policy than the standard MPC implementation.
      \item We demonstrate that our proposed method, incorporating the novel idea of using a surrogate robot model, successfully controls a bipedal inverted pendulum system in real time. 
\end{itemize}

The rest of the paper is organized as follows. Section \ref{sec:related_work} describes related work. Section \ref{sec:preliminaries} provides preliminaries on reinforcement learning and model predictive control methods. Section \ref{sec:method} describes our proposed two-stage learning framework for training a hierarchical control strategy. Section \ref{sec:experiments} provides the experimental settings, and Section \ref{sec:results} presents the experimental results. Section \ref{sec:conclusion} concludes this paper.

%% file: section/2related_works.tex
\subsection{Real-time Robot Control Using Model Predictive Control (MPC)}
Model Predictive Control (MPC) has been widely applied to numerous robotic control problems. However, utilizing a full-body dynamics model in MPC presents challenges for high-dimensional, multi-degree-of-freedom systems or control tasks in complex environments due to significant computational demands. Most real-world applications of MPC employing a complete system dynamics model have been limited to low-dimensional systems, such as remote-controlled cars or hexacopters, in control tasks \citep{williams2016aggressive, neunert2016fast}. Consequently, applications of full-body MPC to high-dimensional, multi-degree-of-freedom robotic systems, such as humanoid robots, are predominantly confined to simulations \citep{erez2013integrated, tassa2012synthesis}, with many practical implementations simplifying the system model \citep{sleiman2021unified, farshidian2017efficient, SRBDMPC, humanoid_centoroidal}. Such approximation methods have proven effective, especially when the limbs' weight is sufficiently small compared to the total body weight, enabling the generation of dynamic motions. Alternatively, some studies have applied full-body MPC to humanoid robots by separating the robot's dynamics into fast and slow dynamics, optimizing the fast dynamics at a finer resolution \citep{ishihara2019full}. By focusing on the structure of the optimal control problem, a quadruped robot successfully achieved jumping motions using MPC that includes optimization of switch times with a full-body dynamics model \citep{katayama2022whole}. However, the MPC used in these studies requires a differentiable dynamics model, necessitating careful design at discontinuities in dynamics or rewards compared to probabilistic sampling-based methods \citep{botev2013cross, goschin2013cross, hansen2003reducing}. Additionally, there is research on solving optimal control problems in advance and utilizing the precomputed solutions in real time, known as Explicit MPC \citep{ExplicitMPC, Explicitblend, vs_explict_mpc}, which reduces computation costs to a level feasible for embedded controllers due to its minimal online computational demands, though its application is mainly limited to linear systems. Efforts to integrate machine learning frameworks to reduce MPC's computational cost have been increasingly pursued. In particular, learning the terminal value of MPC from data helps shorten the required prediction horizon and reduce computation time \citep{polo, mpq, mpqlam}. However, this terminal value depends on the learned task, which does not allow for changes in control objectives during task execution.

\subsection{Goal-Conditioned Reinforcement Learning for Robot Control}
Reinforcement learning methods typically train an agent to achieve a single goal, whereas Goal-Conditioned Reinforcement Learning (GCRL) \citep{liu2022goal} enables an agent to learn to accomplish multiple objectives. GCRL has been widely adopted in robot control studies. For example, a robot's target speed or position can be specified as goal variables, and the learning process is designed to fulfill these objectives \citep{zhu2021mapgo, florensa2018automatic, tang2021hindsight}. Using images as goal variables offers a more direct form of representation, which is frequently employed in complex tasks. Algorithms that utilize images as goal variables have demonstrated effectiveness in several simulations \citep{wardefarley2018unsupervised, lee2020weaklysupervised, chanesane2021goalconditioned}, and their application to 7-degree-of-freedom real robotic arms has also been reported \citep{nair2018visual, nair2019contextual}. In these cases, the goal variable images represent target locations and other objectives.

%% file: section/3preliminaries.tex
\subsection{Reinforcement Learning} \label{subsec:rl}
In reinforcement learning, an agent learns policies through interactions with the environment represented as a Markov Decision Process (MDP). An MDP is defined by a tuple $\mathcal{M} = (\mathcal{X}, \mathcal{U}, \mathcal{F}, \mathcal{C}, \gamma)$, where $\mathcal{X}$ and $\mathcal{U}$ represent the continuous state and action spaces, respectively. $\mathcal{F}$ represents the state transition probability function, and $\mathcal{C}$ represents the cost function at each step, while $\gamma$ denotes the discount factor. At time $t$, the agent's state is denoted as $\boldsymbol{x}_t \in \mathcal{X}$, and the action is denoted as $\boldsymbol{u}_t \in \mathcal{U}$. The action is often defined by control inputs to the robot, such as joint angle commands, in the context of robot control. The transition probability to the next state $\boldsymbol{x}_{t+1}$ after taking action $\boldsymbol{u}_t$ in state $\boldsymbol{x}_t$ is defined as $\boldsymbol{f}(\boldsymbol{x}_{t+1}| \boldsymbol{x}_t, \boldsymbol{u}_t)$. Furthermore, the cost function $c(\boldsymbol{x}_t, \boldsymbol{u}_t)$ is defined as an indicator of the quality of the agent's state and action at time $t$. 

The probabilistic decision-making process of the agent based on its state is referred to as a policy, denoted as $\pi (\boldsymbol{u}_t| \boldsymbol{x}_t)$. The policy indicates the probability of taking action $\boldsymbol{u}_t$ in state $\boldsymbol{x}_t$. 
The goal of reinforcement learning is to find the optimal policy $\pi^*$ that minimizes the expected value of cumulative cost, which can be formulated using a finite number of steps $T$ as follows:
\begin{equation}
\pi^* = \argmin_{\pi} \mathbb{E}_{\pi} \left[\sum_{t=0}^{T} c(\boldsymbol{x}_t, \boldsymbol{u}_t) | \boldsymbol{x}_0 = \boldsymbol{x} \right].
\end{equation}
When the steps continue indefinitely, to prevent the divergence of cumulative cost, a discount rate $\gamma \in [0,1)$ is introduced, expressed as:
\begin{equation} \label{eq:rl_fund}
\pi^* = \argmin_{\pi} \mathbb{E}_{\pi} \left[\sum_{t=0}^{\infty} \gamma^t c(\boldsymbol{x}_t, \boldsymbol{u}_t) | \boldsymbol{x}_0 = \boldsymbol{x} \right].
\end{equation}
A small discount rate $\gamma$ near $0$ encourages short-sighted actions, while a value closer to 1 seeks to minimize the total cost over the long term. Under policy $\pi$, the sum of the discounted cumulative cost expected to be incurred when the agent's state is $\boldsymbol{x}$ is called the state value $V^{\pi}(\boldsymbol{x})$, defined as follows:
\begin{equation}
V^{\pi}(\boldsymbol{x}) := \mathbb{E}_{\pi} \left[\sum_{t=0}^{\infty} \gamma^t c(\boldsymbol{x}_t, \boldsymbol{u}_t) | \boldsymbol{x}_0 = \boldsymbol{x} \right].
\end{equation}

\subsection{Value Function Approximation} \label{approximation}
The optimal value function is defined as the expected discounted cumulative cost that an agent receives under the optimal policy. The Bellman operator $\mathcal{B}$ is defined in Eq.\eqref{eq:belman}, and by applying it iteratively to any bounded function, the optimal value function can be determined \citep{sutton}. The optimal value function $V^*$ is the fixed point of the Bellman operator $\mathcal{B}$, as shown in Eq.\eqref{eq:fixed_point}, where $\boldsymbol{x^\prime}$ denotes the next state of $\boldsymbol{x}$.
\begin{equation} \label{eq:belman}
\mathcal{B} V(\boldsymbol{x}) := \min_{\boldsymbol{u}} \mathbb{E} \left[c(\boldsymbol{x}, \boldsymbol{u}) + \gamma V(\boldsymbol{x}^\prime) \right].
\end{equation}
\begin{equation} \label{eq:fixed_point}
V^*(\boldsymbol{x}) = \mathcal{B} V^*(\boldsymbol{x}) \hspace{5pt} \forall \boldsymbol{x} \in \mathcal{X}.
\end{equation}
Dynamic programming methods, such as value iteration \citep{dp}, can be used to find the optimal value function, and the optimal policy can be described as follows:
\begin{equation}
\pi(\boldsymbol{x}) = \argmin_{\boldsymbol{u}} \mathbb{E} \left[c(\boldsymbol{x}, \boldsymbol{u}) + \gamma V(\boldsymbol{x}^\prime) \right].
\end{equation}
However, accurately computing the optimal value function in continuous MDPs is generally challenging, except in specific cases such as Linear Quadratic Regulators (LQR) \citep{lqr}. Consequently, various methods for approximating the value function have been proposed, with fitted value iteration being one of the standard approaches \citep{bertsekas1996neuro, munos2008finite}. In this approach, function approximators such as neural networks are used to approximate the optimal value function. The approximated value function is parameterized by $\boldsymbol{\theta}$ as $\hat{V}_{\boldsymbol{\theta}}(\boldsymbol{x})$, and $\boldsymbol{\theta}$ is updated as follows:
\begin{equation} \label{eq:theta_update}
\boldsymbol{\theta} \leftarrow \argmin_{\boldsymbol{\theta}} \mathbb{E}_{\boldsymbol{x} \sim \nu} \left[\left(\hat{V}_{\boldsymbol{\theta}}(\boldsymbol{x}) - y(\boldsymbol{x}) \right)^2 \right].
\end{equation}
Here, $\nu$ represents the sampling distribution of the state set, and the target $y$ is computed as $y(\boldsymbol{x}) = \min_{\boldsymbol{u}} \mathbb{E} \left[c(\boldsymbol{x}, \boldsymbol{u}) + \gamma \hat{V}_{\boldsymbol{\theta}}(\boldsymbol{x}^\prime) \right]$. The policy is then obtained as follows:
\begin{equation}
\pi(\boldsymbol{x}) = \argmin_{\boldsymbol{u}} \mathbb{E} \left[c(\boldsymbol{x}, \boldsymbol{u}) + \gamma \hat{V}_{\boldsymbol{\theta}}(\boldsymbol{x}^\prime) \right].
\end{equation}

\subsection{Model Predictive Control (MPC)} \label{subsec:mpc}
In Model Predictive Control (MPC), at each time step, a locally optimal control sequence for a finite future horizon $H$ is computed based on predictions from the nonlinear dynamics model $\boldsymbol{f}$. The problem solved at time step $k$ is formulated as follows, known as the optimal control problem:
\begin{equation} \label{eq:min_mpc}
\begin{aligned}
&\boldsymbol{U}_k^* = \argmin_{\boldsymbol{U}_k} \sum_{t=k}^{k+H-1} c(\boldsymbol{x}_t, \boldsymbol{u}_t) + \phi(\boldsymbol{x}_{k+H}), \\
&\text{s.t.} \quad \boldsymbol{x}_{t+1} = \boldsymbol{f}(\boldsymbol{x}_t, \boldsymbol{u}_t).
\end{aligned}
\end{equation}
where $\boldsymbol{U}_k^* \equiv (\boldsymbol{u}_k^*, \boldsymbol{u}_{k+1}^*, \ldots ,\boldsymbol{u}_{k+H-1}^*)$ represents the computed optimal control sequence. The term $\phi(\boldsymbol{x}_{k+H})$ is referred to as the terminal cost or terminal value, representing the cost at the terminal state $\boldsymbol{x}_{k+H}$. Upon obtaining the optimal control sequence $\boldsymbol{U}_k^*$ at step $k$, the initial value $\boldsymbol{u}_k^*$ is applied to the actual system. Following the observation of the new state at step $k+1$, a new optimal control sequence $\boldsymbol{U}_{k+1}^*$ is computed. This process is repeated at each step, thereby implementing a state feedback control method that depends on the current state. When applying MPC to real systems like robots, the optimal control sequence must be computed in real-time, meaning that Eq.\eqref{eq:min_mpc} needs to be solved within each control period.
The prediction horizon $H$ of MPC is a crucial parameter for real-time control. Under the assumption of an accurate system model, a larger $H$ leads to better solutions but also increases the computational cost. Furthermore, the terminal value $\phi(\boldsymbol{x}_{k+H})$ is an important factor contributing to the performance of MPC. Since MPC only takes a finite future into account to obtain the optimal control sequence, treating the terminal cost as the value at the terminal state allows for the substitution of predictions beyond $\boldsymbol{x}_{k+H}$. By introducing a discount rate $\gamma$ into Eq.\eqref{eq:min_mpc}, the optimized control sequence can be expressed as follows:
\begin{equation} \label{eq:gamma_min_mpc}
\begin{aligned}
&\boldsymbol{U}_k^* = \argmin_{\boldsymbol{U}_k} \sum_{t=k}^{k+H-1} \gamma^t c(\boldsymbol{x}_t, \boldsymbol{u}_t) + \gamma^H \phi(\boldsymbol{x}_{k+H}), \\
&\text{s.t.} \quad \boldsymbol{x}_{t+1} = \boldsymbol{f}(\boldsymbol{x}_t, \boldsymbol{u}_t).
\end{aligned}
\end{equation}
This representation, incorporating a discount rate, considers future uncertainties and has been verified as an effective formulation \citep{granzotto2020finite}. Eq.\eqref{eq:gamma_min_mpc} modifies the optimal policy from an infinite to a finite future horizon as presented in Eq.\eqref{eq:rl_fund}, with its terminal value analogous to the value function in reinforcement learning. Hence, a better approximation of the terminal value enables MPC to achieve better control outputs \citep{polo, mpq, mpqlam}. 

To solve Eq.\eqref{eq:min_mpc}, we need to rely on numerical methods. In this study, we use MPPI \citep{williams2018information} as the MPC algorithm.

%% file: section/4method.tex
\begin{figure*}[h] \centering \includegraphics[width=\linewidth]{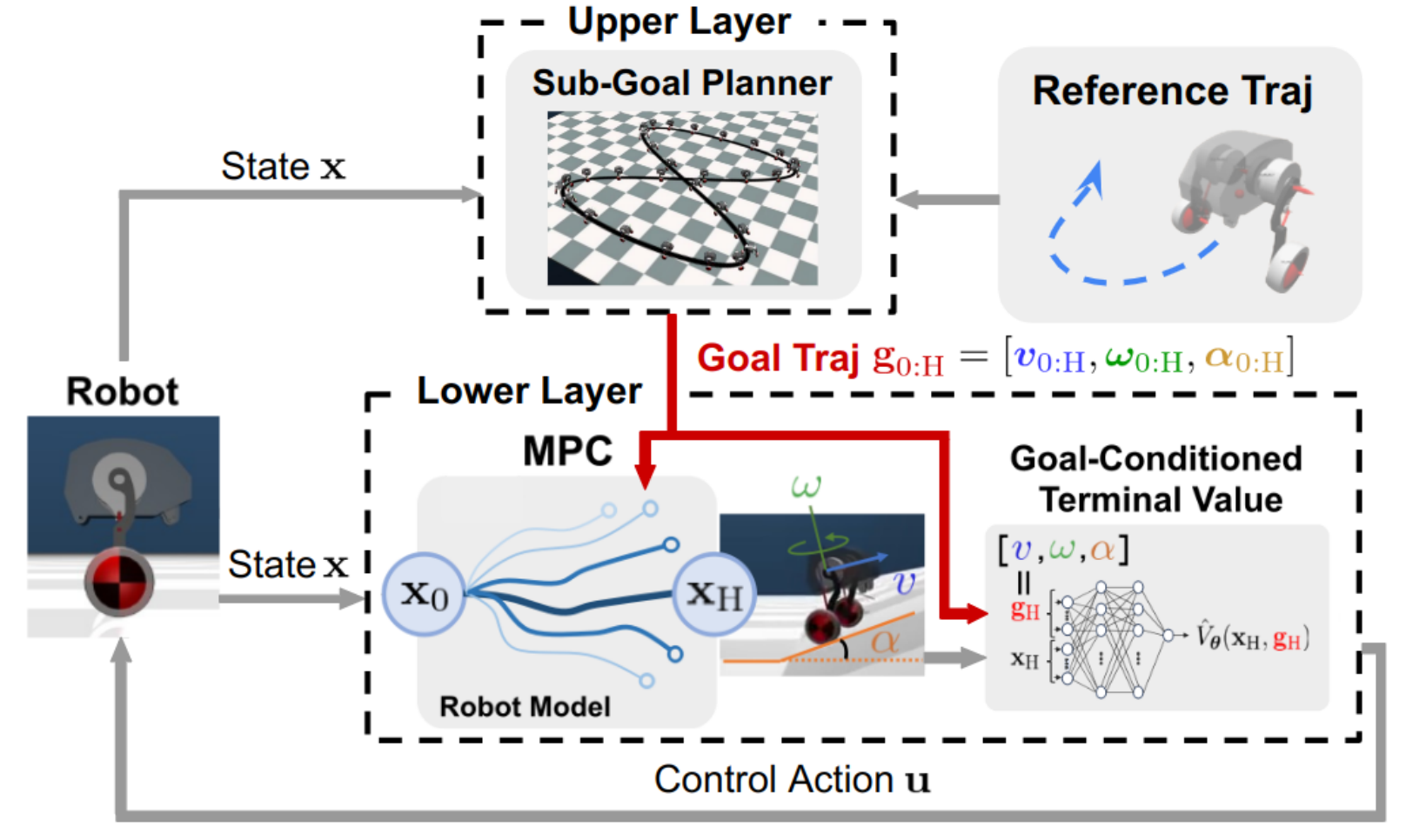} \caption{Concrete implementation to control the bipedal inverted pendulum system. Upper layer: During training, it outputs random goal variables. During inference, it generates a sequence of goal variables, $\boldsymbol{g}_{0
}$, of the same length as the prediction horizon, tailored to the robot's state and reference trajectory (e.g., desired robot position and velocity). Lower layer: learns terminal values corresponding to the goal variables. During inference, it optimizes the predicted trajectory and generates actions in accordance with the commands received from the upper layer, using the learned terminal values. $\boldsymbol{x_t}$, $\boldsymbol{u_t}$, $v$, $\omega$, and $\alpha$ represent the state, control input, robot's linear velocity, angular velocity, and slope angle, respectively.} \label{fig:proposed_method}
\end{figure*}

Here, we present a hierarchical control approach, structured as shown in Fig.\ref{fig:proposed_method}. The framework of the proposed method consists of lower and upper layers.

\subsection{Lower Layer}
The lower layer focuses on learning goal-conditioned terminal values, which are then utilized by MPC to generate control sequences according to given objectives. Recent studies have introduced methods for approximating terminal values from data \citep{polo, mpq, mpqlam}. However, learning terminal values in MPC tends to specialize in a single task. Therefore, we propose learning goal-conditioned terminal values, $\hat{V}_{\theta}(\boldsymbol{x}, \boldsymbol{g})$, where $\boldsymbol{g}$ represents the goal variables.
\begin{equation} \label{eq:train_value}
\begin{aligned}
&y(\boldsymbol{x}, \boldsymbol{g}) = \\
&\min_{\boldsymbol{u}_{0:H-1}} \mathbb{E}\left[\sum_{t=0}^{H-1} \gamma^t c(\boldsymbol{x}_t, \boldsymbol{u}_t, \boldsymbol{g}_t) + \gamma^H \hat{V}_{\boldsymbol{\theta}}(\boldsymbol{x}_H, \boldsymbol{g}_H) | \boldsymbol{x}_0 = \boldsymbol{x} \right].
\end{aligned}
\end{equation}
\begin{equation} \label{eq:train_update}
\boldsymbol{\theta} \leftarrow \argmin_{\boldsymbol{\theta}} \mathbb{E}_{\boldsymbol{x}, \boldsymbol{g} \sim \nu} \left [\left(\hat{V}_{\boldsymbol{\theta}}(\boldsymbol{x}, \boldsymbol{g}) - y(\boldsymbol{x}, \boldsymbol{g})\right)^2 \right].
\end{equation}
Here, $y(\boldsymbol{x}, \boldsymbol{g})$ represents the target value for the terminal values at state $\boldsymbol{x}$ and the goal variable $\boldsymbol{g}$. This process utilizes the concept of Goal-Conditioned Reinforcement Learning (GCRL) \citep{liu2022goal}, where a single cost or reward function is typically defined for each task, leading to learned values that are specialized for the specific goal. Therefore, by setting goal variables separately from the state, as shown in Eq.\eqref{eq:gcrl}, it is possible to train agents capable of achieving a variety of tasks, a process referred to as GCRL.
\begin{equation} \label{eq:gcrl}
\pi^* = \argmin_{\pi} \mathbb{E}_{\pi} \left[\sum_{t=0}^{\infty} \gamma^t c(\boldsymbol{x}_t, \boldsymbol{u}_t, \boldsymbol{g}) | \boldsymbol{x}_0 = \boldsymbol{x} \right].
\end{equation}
The configuration of the goal variables $\boldsymbol{g}$ varies depending on the task.

In Eq.\eqref{eq:train_value}, the target values for the value function were calculated using the solution of the optimal control problem. Using the terminal values obtained from Eq.\eqref{eq:train_value} and Eq.\eqref{eq:train_update}, the MPC can effectively operate as a longer-horizon controller, with the flexibility to modify control objectives by varying the goal variables. During training, the goal variable $\boldsymbol{g}$ remains constant throughout the horizon and is randomly sampled in each episode. During the inference phase, however, the goal variable is instead represented as a time-varying sequence. The key distinction from the standard GCRL framework is that the goal variable forms a trajectory over the horizon during inference, which helps smooth the agent's motions.

\subsection{Upper Layer}
The upper layer generates goal variables that serve as inputs to the terminal value function for the lower-layer MPC. By varying objectives according to the robot's state, the controller in the lower layer can achieve flexible and diverse motion generation suited to the given environment. The role of the trajectory generator in the upper layer differs between the training and inference phases. In the training phase, the upper layer generates random goal variables within a certain range to ensure the lower layer's terminal value can accommodate a wide range of objective variables. In the inference phase, to enable the robot to generate diverse motions suited to the given environment, a feedback goal variable generator is required, which takes the observed robot's state as input and outputs goal variables. Specifically, a sequence of goal variables corresponding to the number of steps in MPC's prediction horizon is generated each control cycle. The key aspect of the method is using a time-varying sequence of goal variables rather than a time-invariant one, which leads to smoother motion generation for the robot. The way of generating goal variable sequences is task-dependent and can be implemented using either model-based or data-driven approaches. The proposed algorithm is summarized in Algorithms \ref{proposed_training} and \ref{proposed_inference}.

\begin{algorithm}
    \caption{Proposed Method (Training)}
    \label{proposed_training}
    \begin{algorithmic}[1]
    \renewcommand{\algorithmicrequire}{\textbf{Given:}}
    \REQUIRE $H$: Planning Horizon, $\boldsymbol{\theta}$: Value function parameters, $n$: mini-batch size, \\
    $G$: number of gradient steps, $Z$: update frequency,  \\
    $\boldsymbol{g_{\text{min}}}, \boldsymbol{g_{\text{max}}}$: minimum and maximum of goal variables;
    \FOR{$t \leftarrow 0 \,\, \textbf{to} \,\, \infty$}
     \STATE Sample $\boldsymbol{g}_{\text{target}}$ from a continuous uniform distribution $U(\boldsymbol{g_{\text{min}}}, \boldsymbol{g_{\text{max}}})$;
     \STATE Determine control input $\boldsymbol{u}_t$ according to MPC Eq.\eqref{eq:gamma_min_mpc} with estimated terminal value $\hat{V}_{\boldsymbol{\theta}}(\boldsymbol{x}_{t+H}, \boldsymbol{g}_{\text{target}})$;
     \STATE Execute $\boldsymbol{u}_t$ on the system and add state experience $\boldsymbol{x}_t$ to the replay buffer $\mathcal{D}$;
     \IF{$t\%Z==0$}
        \FOR{$G$ times}
            \STATE Sample $n$ states from the replay buffer and compute targets using Eq.\eqref{eq:train_value};
            \STATE Update the terminal value parameters using Eq.\eqref{eq:train_update};
        \ENDFOR
     \ENDIF
    \ENDFOR
    \end{algorithmic} 
\end{algorithm}

\begin{algorithm}
    \caption{Proposed Method (Inference)}
    \label{proposed_inference}
    \begin{algorithmic}[1]
    \renewcommand{\algorithmicrequire}{\textbf{Given:}}
    \REQUIRE $H$: Planning Horizon, $G_{\mathrm{plan}}$: Sub-Goal Planner, $\boldsymbol{\theta}$: Value function parameters;
    \WHILE{\textit{task not completed}}
     \STATE Observe current state $\boldsymbol{x}_t$ and update MPC's internal state;
     \STATE Sample planned goal variable trajectory $\boldsymbol{g}^{\text{plan}}_{t:t+H} \sim G_{\mathrm{plan}}(\boldsymbol{x}_t)$;
     \STATE Determine control input $\boldsymbol{u}_t$ according to MPC Eq.\eqref{eq:gamma_min_mpc} with estimated terminal value $\hat{V}_{\theta}(\boldsymbol{x}_{t+H}, \boldsymbol{g}^{\text{plan}}_{t+H})$;
     \STATE Send $\boldsymbol{u}_t$ to the actuators;
     \STATE Check for task completion;
    \ENDWHILE
    \end{algorithmic} 
\end{algorithm}

\subsection{Domain Randomization for Robust Value Function Estimation}
To achieve robust terminal value estimation, we adopt the Domain Randomization (DR) approach. Commonly used in reinforcement learning, this method involves randomizing the simulation environment to train policies under various conditions. This helps in acquiring robust policies that can adapt to real-world scenarios during testing. The DR approach is widely applied, with randomization targets varying depending on the task. For instance, in manipulation tasks, modifying object coordinates, shapes, colors, and lighting conditions has resulted in robust policies that successfully transfer to real-world scenarios \citep{tobin2017domain}. Additionally, randomizing the robot's dynamics model—such as varying mass or motor control gains—has proven successful in real-world experiments \citep{peng2018sim}. Skills for cube manipulation have been developed using a robotic hand by randomizing both visual information and physical dynamics \citep{andrychowicz2020learning}. In this study, we randomize the terrain to ensure that the optimized policy becomes robust.

\subsection{Faster MPC Calculation with a Surrogate Robot Model}
Predicting the future state sequence of the robot is necessary for deriving the optimal controller in MPC, and faster numerical integration in the physical robot simulation is desirable to achieve this. We propose using a surrogate robot model instead of the actual robot model to accomplish this. The surrogate robot model has the same degrees of freedom as the actual robot system. This approach is advantageous because some mechanical structures, such as parallel link mechanisms, slow down numerical integration due to the need to satisfy closed-loop mechanical constraints. Specifically, in this study, we use a prismatic joint system as the surrogate model for a parallel-link legged mechanism to enable faster MPC calculations.

%% file: section/5experiments.tex
\subsection{The Bipedal Inverted Pendulum Robot}
\begin{figure}[ht]
    \centering
    \begin{minipage}[b]{0.5\linewidth}
        \centering
        \subfloat[Bipedal Inverted Pendulum Robot]{\includegraphics[width=0.7\linewidth]{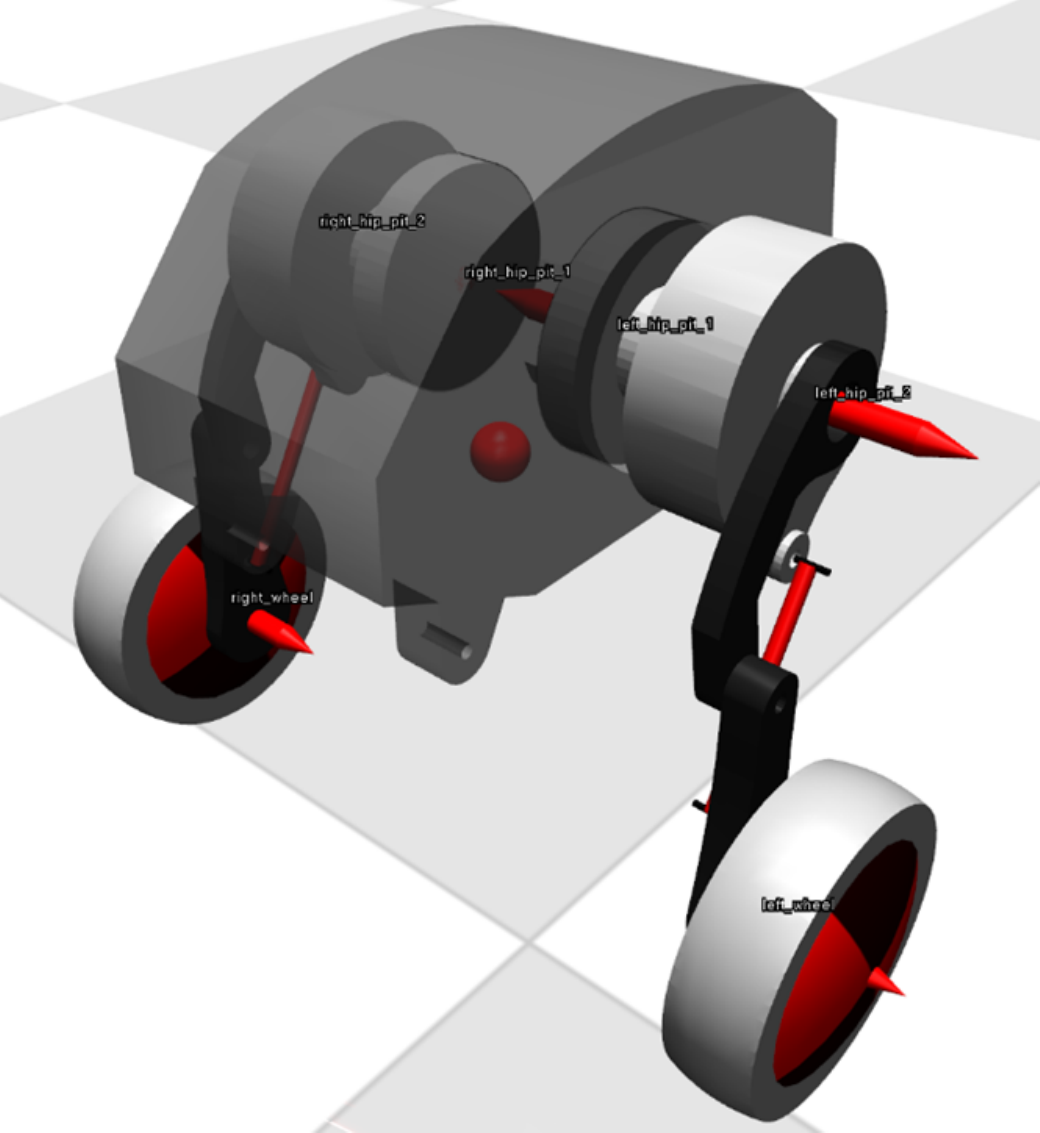}}
    \end{minipage}
    \hspace{0.5cm} 
    \begin{minipage}[b]{0.4\linewidth}
        \centering
        \subfloat[Control variables]{\includegraphics[width=\linewidth]{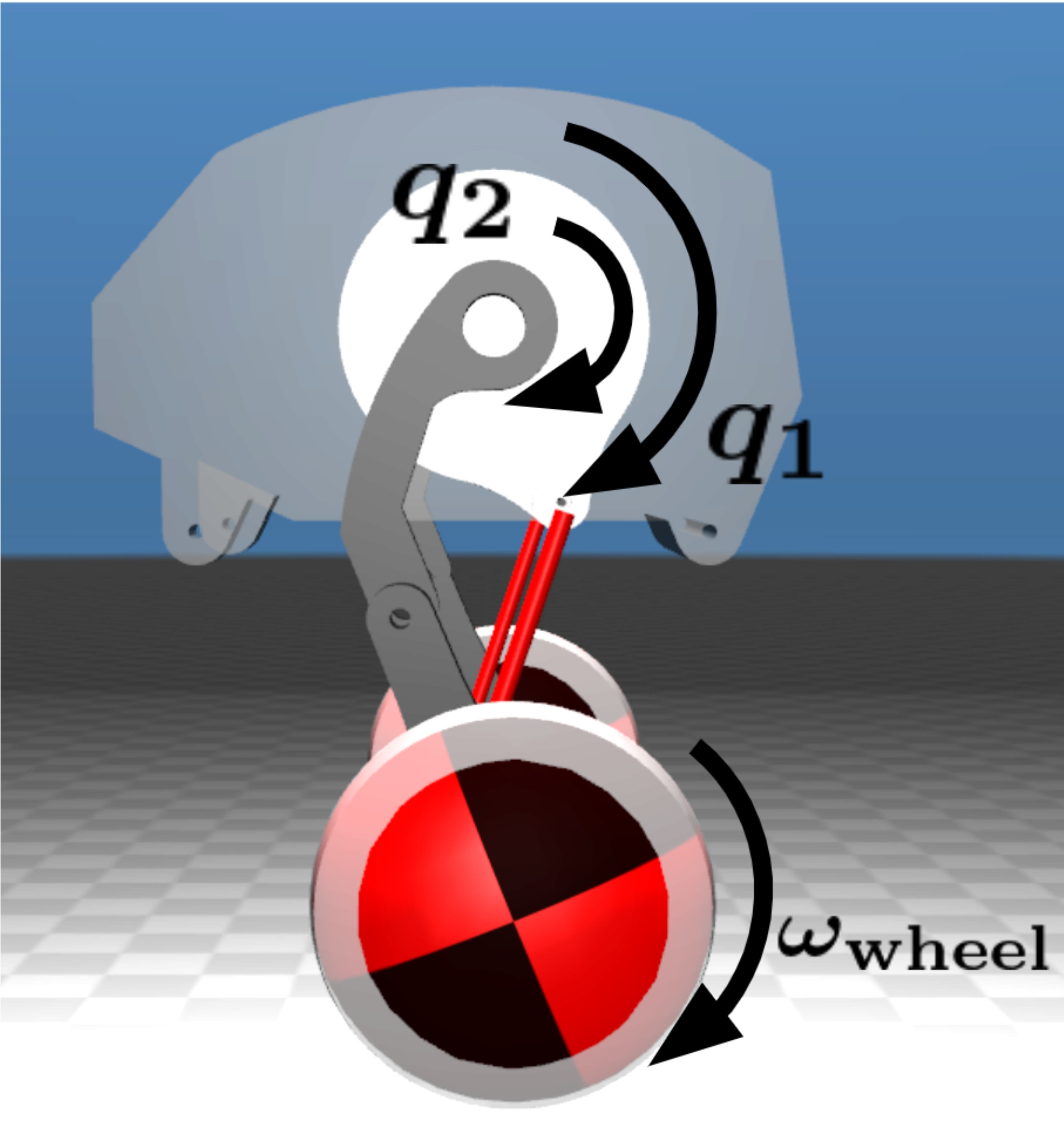}}
    \end{minipage}
     \caption{Simulated robot model. $q_1$ and $q_2$ represent joint angles, and $\omega_{\mathrm{wheel}}$ denotes the angular velocities of the wheels.}
  \label{fig:diablo_sim}
\end{figure}

In this study, we used a simulated model of the bipedal inverted pendulum robot Diablo developed by Direct Drive Technology \citep{liu2024diablo}. This robot model is implemented in the MuJoCo simulation environment \citep{todorov2012mujoco} (see also Fig.\ref{fig:diablo_sim}). The robot model has a state dimension of $33$, and its control variables include joint angle commands to each motor and angular velocity commands for the wheels, which make up $6$ dimensions. In Fig.\ref{fig:diablo_sim}, $q_1$ and $q_2$ represent joint angles, and $\omega_{\mathrm{wheel}}$ denotes the angular velocities of the wheels.

We test the proposed method with locomotion tasks on flat and sloped terrains depicted in Fig.\ref{fig:flat_base}. 
The goal variables are set as $\boldsymbol{g}_t = [v_t, \omega_t]$ for the flat surface and $\boldsymbol{g}_t = [v_t, \omega_t, \alpha_t]$ for the sloped surface, and the network model in the lower layer learns the terminal values based on Eqs. \eqref{eq:train_value} and \eqref{eq:train_update}. Here, $v_t$, $\omega_t$, and $\alpha_t$ represent the robot's velocity, angular velocity, and the angle of the slope the robot is facing at time $t$, respectively. 
During inference, a reference trajectory is provided, and the trajectory generator in the upper layer produces goal variable sequences according to the current robot state. We use the Kanayama Control algorithm \citep{kanayama1990stable}, which is widely used for controlling mobile robots, as the trajectory generator. By using the Kanayama control, a series of goal variable sequences over the horizon $\boldsymbol{g}^{\mathrm{plan}}_{t:t+H} = [v_{t:t+H}, \omega_{t:t+H}]$ for the flat surface and $\boldsymbol{g}^{\mathrm{plan}}_{t:t+H} = [v_{t:t+H}, \omega_{t:t+H}, \alpha_{t:t+H}]$ for the sloped surface are derived according to the reference and actual positions and orientations.

\subsection{Path Tracking Tasks} \label{subsec:flat_task}
\begin{figure}[htbp]
\centering
\subfloat[Flat Surface]{\includegraphics[width=0.45\linewidth]{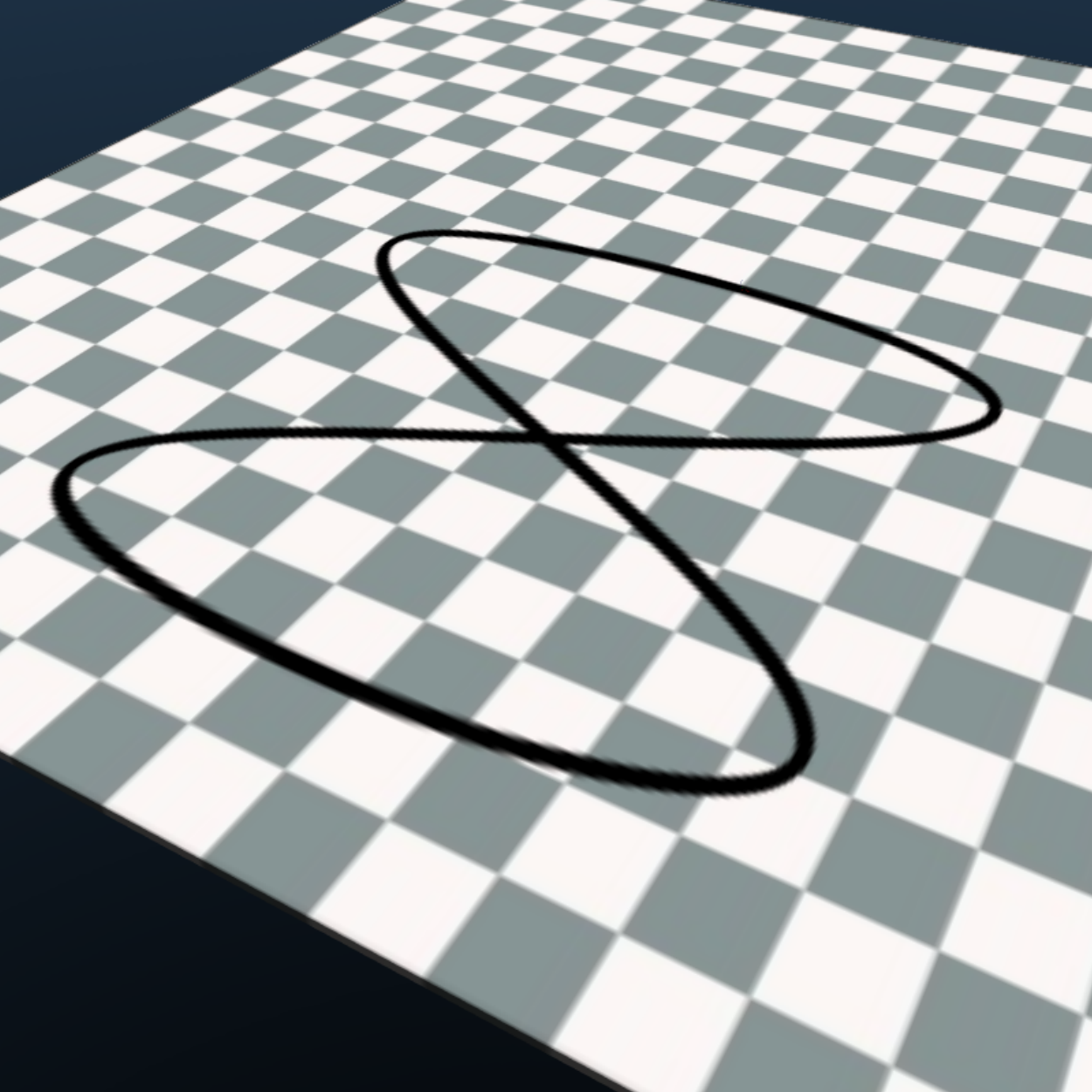}}
\hspace{2mm}
\subfloat[Sloped Surface]{\includegraphics[width=0.45\linewidth]{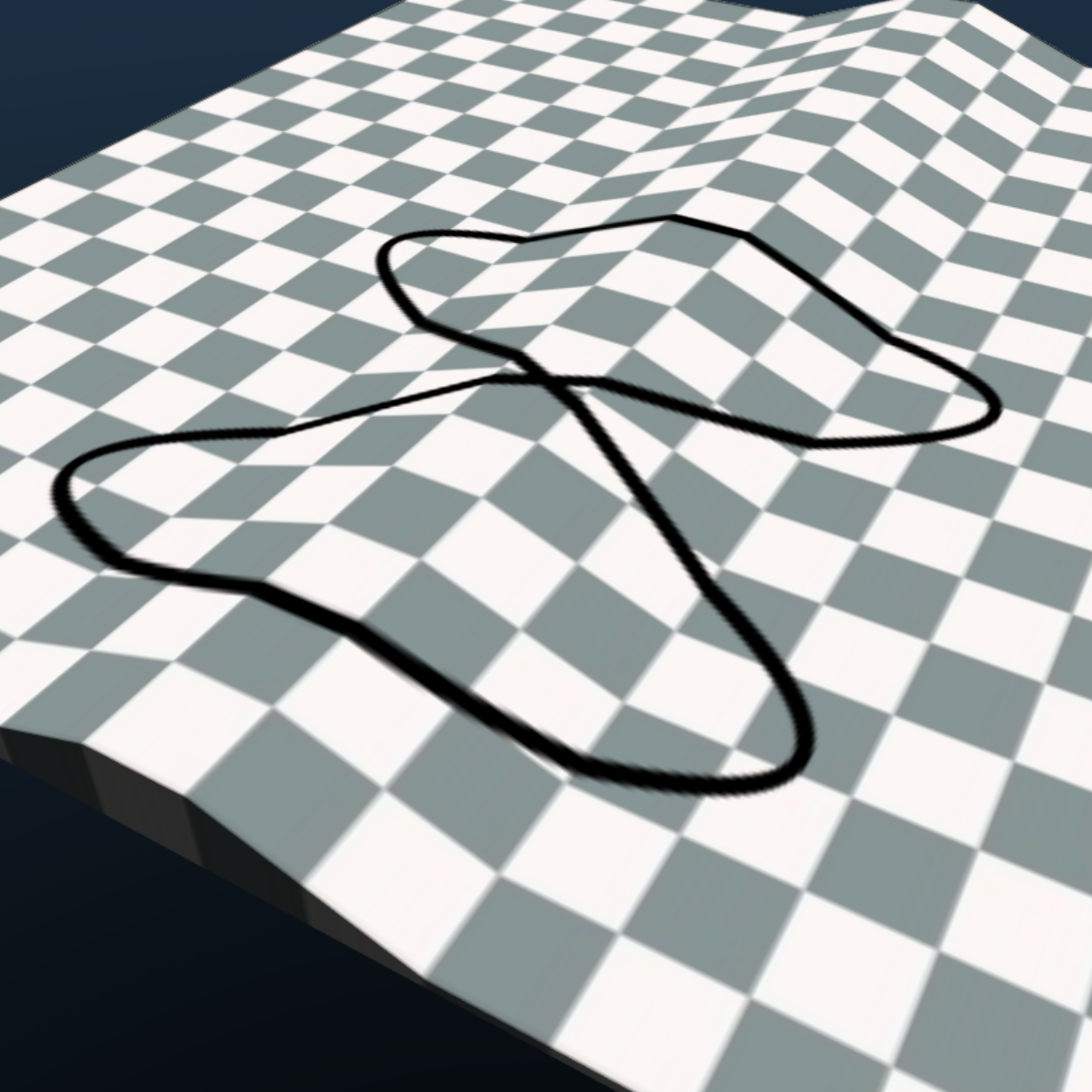}}
\caption{Terrain and lemniscate trajectory for the tracking task.}
\label{fig:flat_base}
\end{figure}

We evaluate the task of following a lemniscate trajectory on flat and sloped surfaces, as illustrated in Fig.\ref{fig:flat_base}.

The lemniscate trajectory is defined by the following equations:
\begin{equation} \label{eq:fig_eight_state}
\begin{aligned}
x_{\mathrm{ref},t} &= r \sin \left(\dfrac{4 \pi t}{T} \right) \\
y_{\mathrm{ref},t} &= r \cos \left(\dfrac{2 \pi t}{T} \right) - r \\
\psi_{\mathrm{ref},t} &= \left. \dfrac{dy(t)}{dx(t)} \right|_{t=t}
\end{aligned}
\end{equation}
Here, $r>0$ determines the size of the figure-eight trajectory, $t$ is the timestep, and $T$ is the total duration of the episode.
The reference velocity and angular velocity trajectories are obtained as follows:
\begin{equation} \label{eq:fig_eight_ctrl}
\begin{aligned}
v_{\mathrm{ref}, t} &= \dfrac{\sqrt{(x_{\mathrm{ref}, t+1} - x_{\mathrm{ref}, t})^2
+ (y_{\mathrm{ref}, t+1} - y_{\mathrm{ref}, t})^2}}{dt} \\
\omega_{\mathrm{ref}, t} &= \dfrac{(\psi_{\mathrm{ref}, t+1} - \psi_{\mathrm{ref}, t})}{dt}
\end{aligned}
\end{equation}
The reference trajectories prepared by Eqs.~\eqref{eq:fig_eight_state} and \eqref{eq:fig_eight_ctrl} are given to Kanayama Control.
Meanwhile, the cost function for the lower-level MPC is set as follows:
\begin{equation} \label{eq:mpc_cost}
\begin{aligned}
c_t &= 10 \, d(5, v_{\mathrm{base},t}, v_{\mathrm{goal},t}) + 5 \, d(5, \omega_{\mathrm{base},t}, \omega_{\mathrm{goal},t}) \\
&+ 3 \, \mathbbm{1}(p_{z_\mathrm{{base,wheel}}} < z_{\mathrm{th}}) + \boldsymbol{u}_t^\mathrm{T} \boldsymbol{\Sigma}^{-1} \boldsymbol{u}_t.
\end{aligned}
\end{equation}
where $d$ is defined as:
\begin{equation}
d(a,b,\beta) = \left(1 - \exp \left(-\beta \|a-b\|^2 \right)\right).
\end{equation}
Its value ranges from $d(a,b,\beta) \in [0,1)$. The function $d$ aims to minimize the distance between $a$ and $b$, with $d$ approaching 0 as the distance decreases, controlled by $\beta > 0$.
The first and second terms in Eq.~\eqref{eq:mpc_cost} aim to minimize the discrepancy at timestep $t$ between the robot's base velocity and angular velocity, $v_{\mathrm{base},t}, \omega_{\mathrm{base},t}$, and the target velocity and angular velocity output by Kanayama Control, $v_{\mathrm{goal},t}, \omega_{\mathrm{goal},t}$.
The third term incurs a cost when the height of the robot's base from the wheels, $p_{z_\mathrm{{base,wheel}}}$, falls below a threshold $z_{\mathrm{th}}$, which acts to prevent the robot from tipping over.
The fourth term represents the control cost. The matrix $\boldsymbol{\Sigma}$ is the covariance of the noise in MPPI. Detailed parameters are provided in \ref{sec:appendix_hyperparameter}.

During training, the goal variables are sampled at the start of each episode within the range of Table \ref{goal_variables}.
For the sloped surface, in each episode, a slope with angle $\alpha_{\mathrm{target}}$ is prepared, and by solving the MPC to follow the target values $v_{\mathrm{target}}, \omega_{\mathrm{target}}$, data is accumulated to learn the terminal values corresponding to each goal variable. Note that $\alpha$ acts as a parameter to vary the environment and is not explicitly incorporated into the cost function.

\begin{table}[hbtp]
\caption{Goal Variables}
\label{goal_variables}
\centering
\begin{tabular}{|l|l|}
\hline
Goal Variables & Value Range\\
\hline \hline
Body linear velocity $v$ & (0,1)\ [\si{m/s}] \\
Body angular velocity $\omega$ & (-1.2, 1.2)\ [\si{rad/s}] \\
Slope angle $\alpha$ &  (-25, 25)\ [\si{deg}] \\
\hline
\end{tabular}
\end{table}

Corresponding data is retained, and the terminal values conditional on the objectives are learned.
Inputs to the terminal value function are extracted from the robot's state, setting the observation, $\boldsymbol{o} = \phi(\boldsymbol{x})$, where $\phi$ is the function for extracting information.
The robot's observation, $\boldsymbol{o}$, is given by 
$\boldsymbol{o} = [p_{z_\mathrm{{base,wheel}}}, \dot{\boldsymbol{p}}_\mathrm{base,wheel}, \boldsymbol{p}_{\mathrm{base}_{\phi, \theta, \psi}}, \dot{\boldsymbol{p}}_{\mathrm{base}_{\phi, \theta, \psi}}, \boldsymbol{q}, \dot{\boldsymbol{q}}, \boldsymbol{\omega}_{\mathrm{wheel}}] \in \mathbb{R}^{20}$.
Here, $\boldsymbol{p}$ represents the robot's position and orientation, $\boldsymbol{q}$ represents the robot's joint angles, and $\boldsymbol{\omega}$ represents the angular velocity.
An episode ends when the robot model tips over or after 10 seconds have elapsed, and the training is conducted over 150 episodes.

Next, hierarchical control is applied using the learned values. The upper-level Kanayama Control generates a sequence of goal variables based on the robot's state over the prediction horizon. It incurs costs to ensure that the robot follows this goal variable trajectory. The final goal variable in the sequence is used as input for the terminal value, replacing the need to predict the future beyond this point.

\begin{figure}[htbp]
 \centering
 \includegraphics[width=0.7\linewidth]{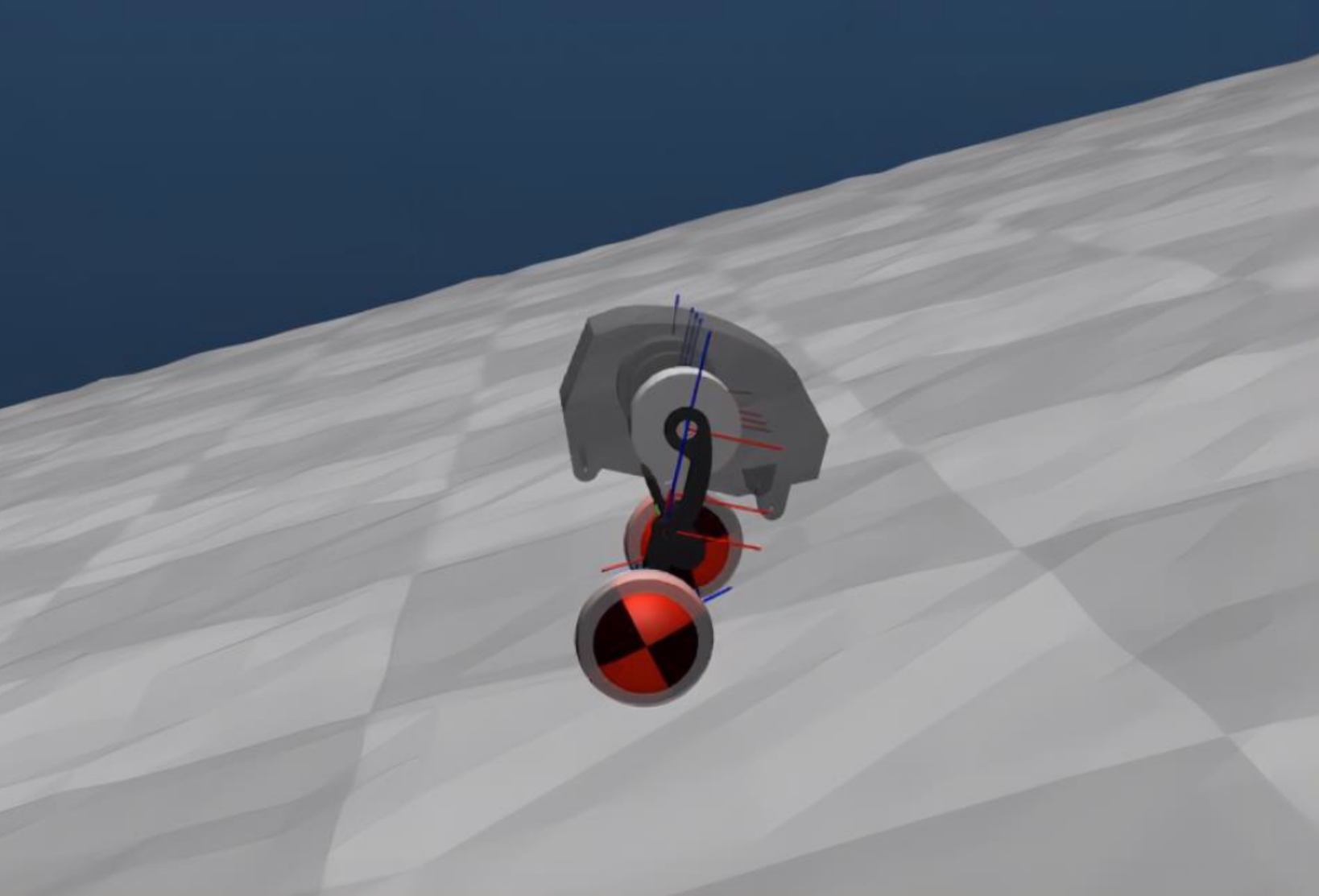}
 \caption{The robot learns to traverse uneven terrain, making the driving module more robust.}
 \label{fig:domain_randomization}
\end{figure}

\paragraph{Domain Randomization}
To improve the robustness of individual terminal values, especially during slope transitions where Diablo's performance deteriorates or tipping occurs, we adapt the approach of environment randomization \citep{tobin2017domain, peng2018sim}, commonly used in reinforcement learning, to aid in learning terminal values. By randomizing the sloped terrain in each episode and making it bumpy with random heights and placements of the floor, it is expected to achieve more robust terminal values, as shown in Fig.\ref{fig:domain_randomization}. In this study, each placement height $z_{\text{terrain}}$ is sampled from $z_{\text{terrain}} \sim U(0, 0.05)\ [\si{m}]$. Therefore, during inference, the first term on the right-hand side of Eq.\eqref{eq:min_mpc} is computed based on a single model's prediction, while the second term represents the terminal value obtained through domain randomization, with each term calculated using different models.

\begin{figure}[htbp]
 \centering
 \includegraphics[width=\linewidth]{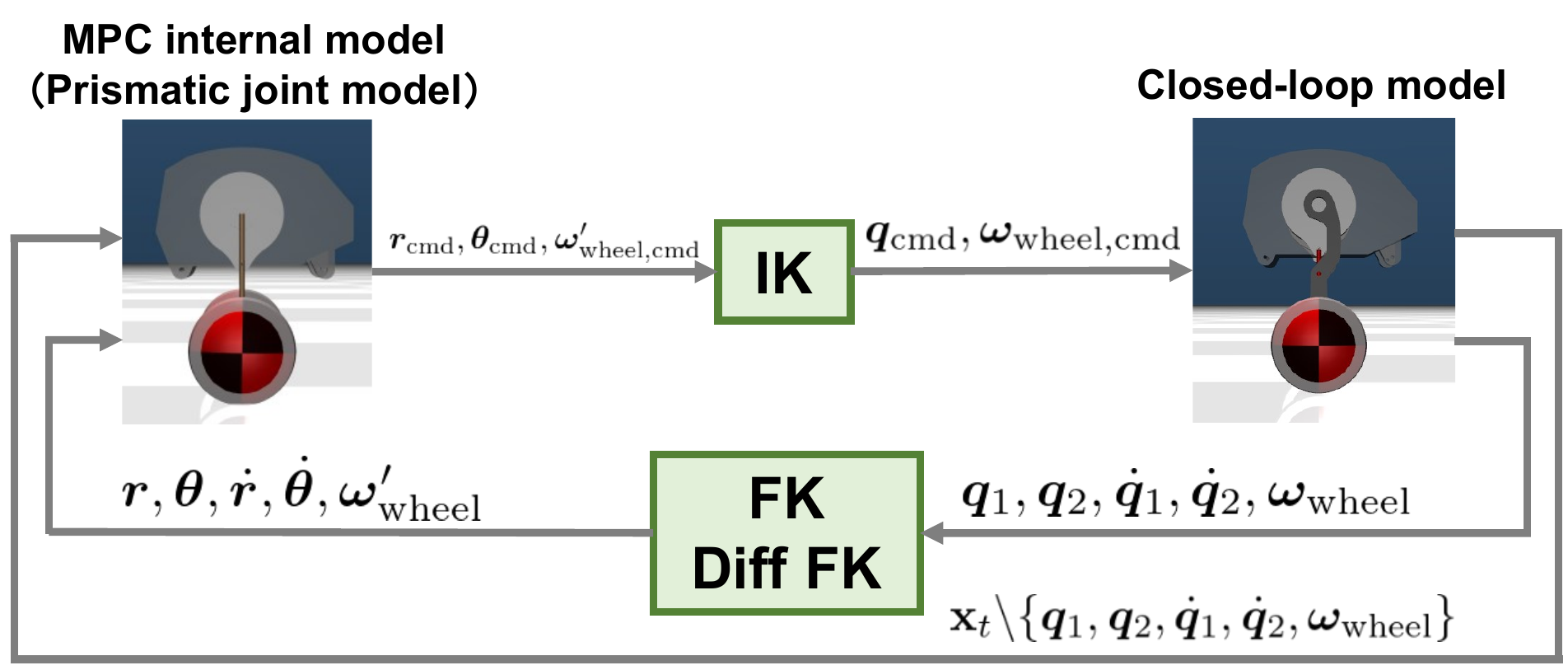}
     \caption{Surrogate Robot Model calculation: In the MPC internal model, a prismatic joint model is used, and it outputs control commands for the length $r_\mathrm{cmd}$, angle $\theta_\mathrm{cmd}$, and the wheel's angular velocity $\omega^\prime_\mathrm{wheel, cmd}$. These commands are then converted into the corresponding joint angles $q_\mathrm{cmd}$ and the wheel's angular velocity $\omega_{\mathrm{wheel, cmd}}$ for the Diablo robot through inverse kinematics and input into the Diablo model. The joint angles $q_1, q_2$, angular velocities $\dot{q}_1, \dot{q}_2$, and the wheel's angular velocity $\omega_\mathrm{wheel}$, as observed from Diablo's closed-loop model, are converted into the prismatic joint model's length $r$, angle $\theta$, and their derivatives $\dot{r}, \dot{\theta}$ through forward kinematics and differential forward kinematics. These are then updated as the state in the MPC's internal model.}
 \label{fig:diablo_loop}
\end{figure}

\paragraph{Surrogate Robot Model Approach}
The dynamic simulation of Diablo's closed-loop link mechanism often encounters instabilities. These instabilities significantly slow down the convergence of solutions. To address this issue, we model the parallel links in the MPC's internal model as linear joints to improve the stability of computations. The transformation between linear joints and parallel links is derived using Forward Kinematics (FK), differential Forward Kinematics, and Inverse Kinematics (IK) \citep{kucuk2006robot}. This process constructs a loop system for the Diablo robot, as depicted in Fig.\ref{fig:diablo_loop}.

\subsection{Computational Setups}
All experiments in this study were conducted on a PC with the following specifications: Intel(R) Core(TM) i9-11900K $3.50$ GHz and $32$ GB RAM. Furthermore, all computations, from the calculation of MPPI to the learning of the terminal value, were performed using C++. In the computation of MPPI, trajectory generation was accelerated through the use of 12-thread multiprocessing. For the training and inference of the neural network, the PyTorch C++ API \citep{Paszke_PyTorch_An_Imperative_2019} was utilized. In all experiments, the control cycle was set at $10$ ms, and the prediction horizon could be represented either in terms of step numbers or time intervals, with the former multiplied by the control cycle to obtain the corresponding time intervals. Details on the hyperparameters of MPPI and the neural network for each task are described in \ref{sec:appendix_hyperparameter}.

%% file: section/6results.tex
\subsection{Path Tracking Tasks} \label{subsec:path_tracking}
\begin{figure*}[htb]
    \centering
    \begin{minipage}[b]{0.5\linewidth}
        \centering
        \includegraphics[width=6.cm]{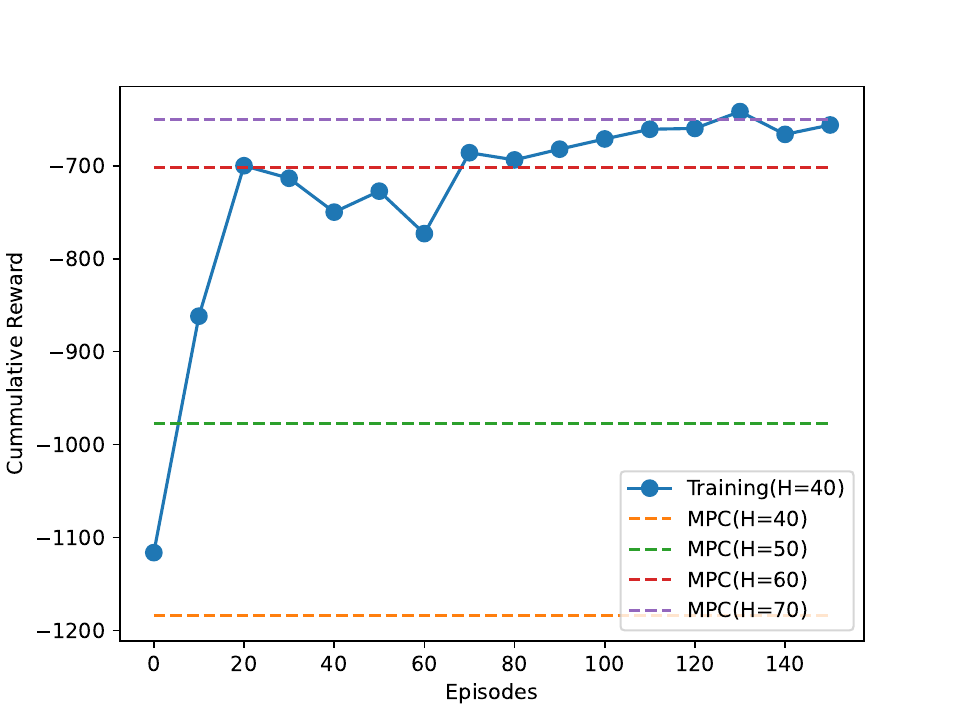}
        \subcaption{Flat surface}
        \label{fig:flat_terrain_training}
    \end{minipage}
    \hspace{5mm}
    \begin{minipage}[b]{0.4\linewidth}
        \centering
        \includegraphics[width=6.5cm]{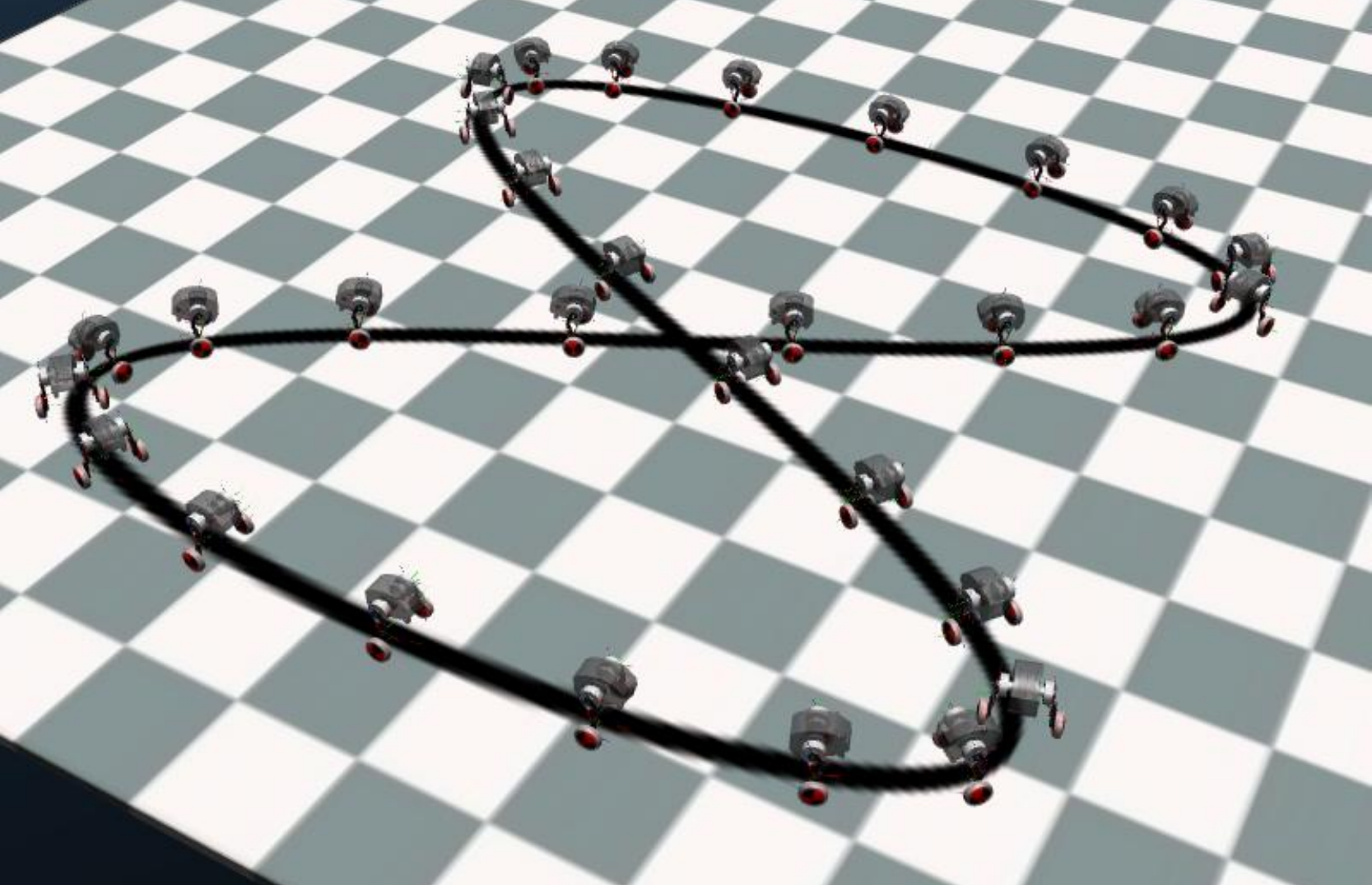}
        \subcaption{Flat surface}
        \label{fig:flat_terrain_traj_track}
    \end{minipage} \\
    \begin{minipage}[b]{0.5\linewidth}
        \centering
        \vspace{-3mm}
        \includegraphics[width=6.cm]{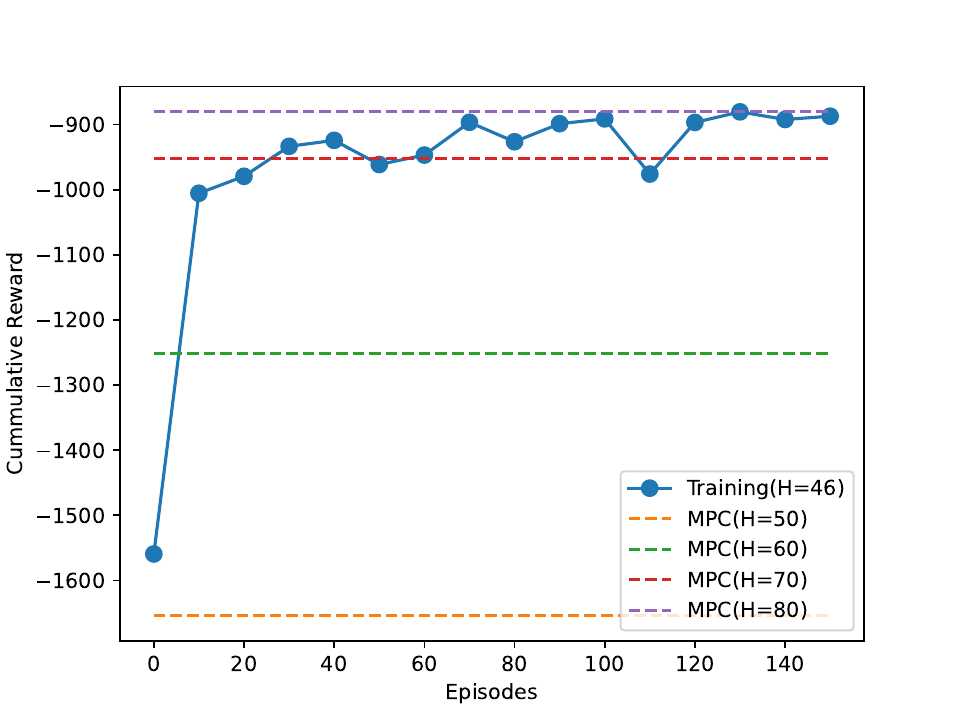}
        \subcaption{Sloped surface}
        \label{fig:slope_terrain_training}
    \end{minipage}
    \hspace{5mm}
    \begin{minipage}[b]{0.4\linewidth}
        \centering
        \includegraphics[width=6.5cm]{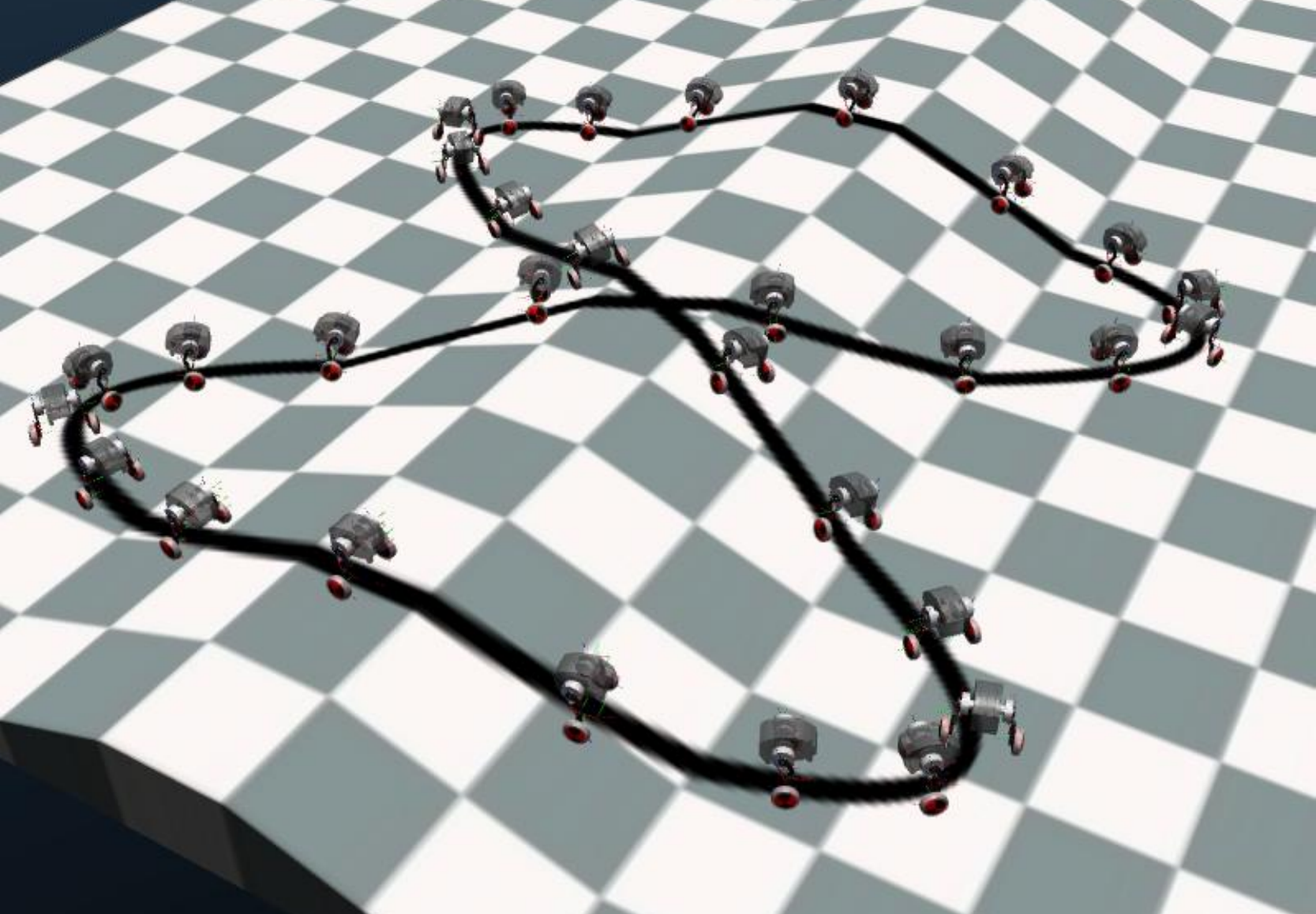}
        \subcaption{Sloped surface}
        \label{fig:slope_terrain_traj_track}
    \end{minipage}
    \caption{Left Figures: The learning curves show cumulative rewards for each episode. (a) depicts the cumulative rewards based on Eq.~\eqref{eq:mpc_cost} when learning terminal value function conditioned on target velocity and angular velocity on flat terrain, while (c) shows rewards for sloped terrain, where the value is also conditioned on the slope angle (higher is better). The dotted lines represent cumulative rewards from MPC without terminal value learning. On flat terrain, horizon lengths $H = 40, 50$ result in the robot falling early in the episode, reducing rewards, while horizons $H=60,70$ enable successful locomotion. For sloped terrain, horizons $H=50,60$ lead to falls, but horizons $H=70,80$ result in successful performance. The solid blue lines show cumulative rewards using the proposed method, where terminal value functions are learned to improve performance. The robot achieves comparable performance to standard MPC with longer horizons $H=70$ (flat) and $H=80$ (sloped), even with shorter horizons $H=40$ (flat) and $H=46$ (sloped).\\
    Right Figures: These figures illustrate the results of using the learned terminal value functions with horizon lengths $H=40$ (flat) and $H=46$ (sloped) to perform lemniscate trajectory tracking. The robot successfully follows the trajectory and adapts to terrain changes. Without terminal value learning, the robot would fall at the beginning of each run with the same horizon lengths.}
    \label{fig:learning_curve}
\end{figure*}

The learning curves obtained under these conditions are shown in the left image of Fig.~\ref{fig:learning_curve}. This image displays the cumulative reward, defined as the negative cumulative cost per episode.

For the evaluation, we set the goal variables as $v_{\mathrm{goal}} = [0.2, 0.4, 0.6, 0.8]$ and $\omega_{\mathrm{goal}} = [-1.0, -0.5, 0, 0.5, 1.0].$ The average cumulative reward for each episode under these settings is illustrated, indicating that the proposed method at $H=40$ significantly outperforms the MPC at $H=40$ and exhibits comparable performance to the MPC at $H=70$, thus successfully learning the terminal values for the specified range of goal variables.

From the figure, it is evident that the motions generated follow the specified figure-eight trajectory. Additionally, after setting the internal model's timestep for the MPC calculations to $5 \, [\si{ms}]$, the maximum computation time for the entire control framework across all timesteps was measured. In 10 trials, the maximum computation time was $7.7 \, [\si{ms}]$, ensuring real-time control within the $10 \, [\si{ms}]$ control cycle.

The learning curves are shown in Fig.~\ref{fig:slope_terrain_training}. For the evaluation, we set the following: $v_{\mathrm{goal}} = [0.2, 0.4, 0.6, 0.8]$, $\omega_{\mathrm{goal}} = [-1.0, -0.5, 0, 0.5, 1.0]$, and $\alpha_{\mathrm{goal}} = [-20, -10, 0, 10, 20].$ Fig.~\ref{fig:slope_terrain_training} represents the average cumulative reward for each episode under these settings, indicating that, through the learning process, the proposed method at $H=46$ exhibits performance comparable to the MPC at $H=80$, thus successfully learning the terminal values for a diverse range of goal variables.
The motions obtained through this process are shown in Fig.~\ref{fig:slope_terrain_traj_track}. The figure displays Diablo at intervals of $2 \ [\si{s}]$.
Fig.~\ref{fig:slope_terrain_traj_track} shows that the motions generated follow the figure-eight trajectory while adjusting for the slope. Similar to the previous section, the maximum computation time for the control framework was measured over 10 episodes, with a maximum computation time of $8.7 \, [\si{ms}]$, ensuring real-time control within the $10 \, [\si{ms}]$ control cycle.

\subsection{Evaluation of Real-Timeness and Control Performance} \label{subsec:realtime}
Section \ref{subsec:path_tracking} demonstrates that the proposed method can generate a variety of motions. In this section, we compare the real-timeness and control performance of the proposed method against several alternative methods. The task involves evaluating the motion of ascending and descending a slope while maintaining constant translational and angular velocities on terrain, as shown in Fig.~\ref{fig:flat_base}b.

\begin{figure}[h]
\centering
\includegraphics[width=\linewidth]{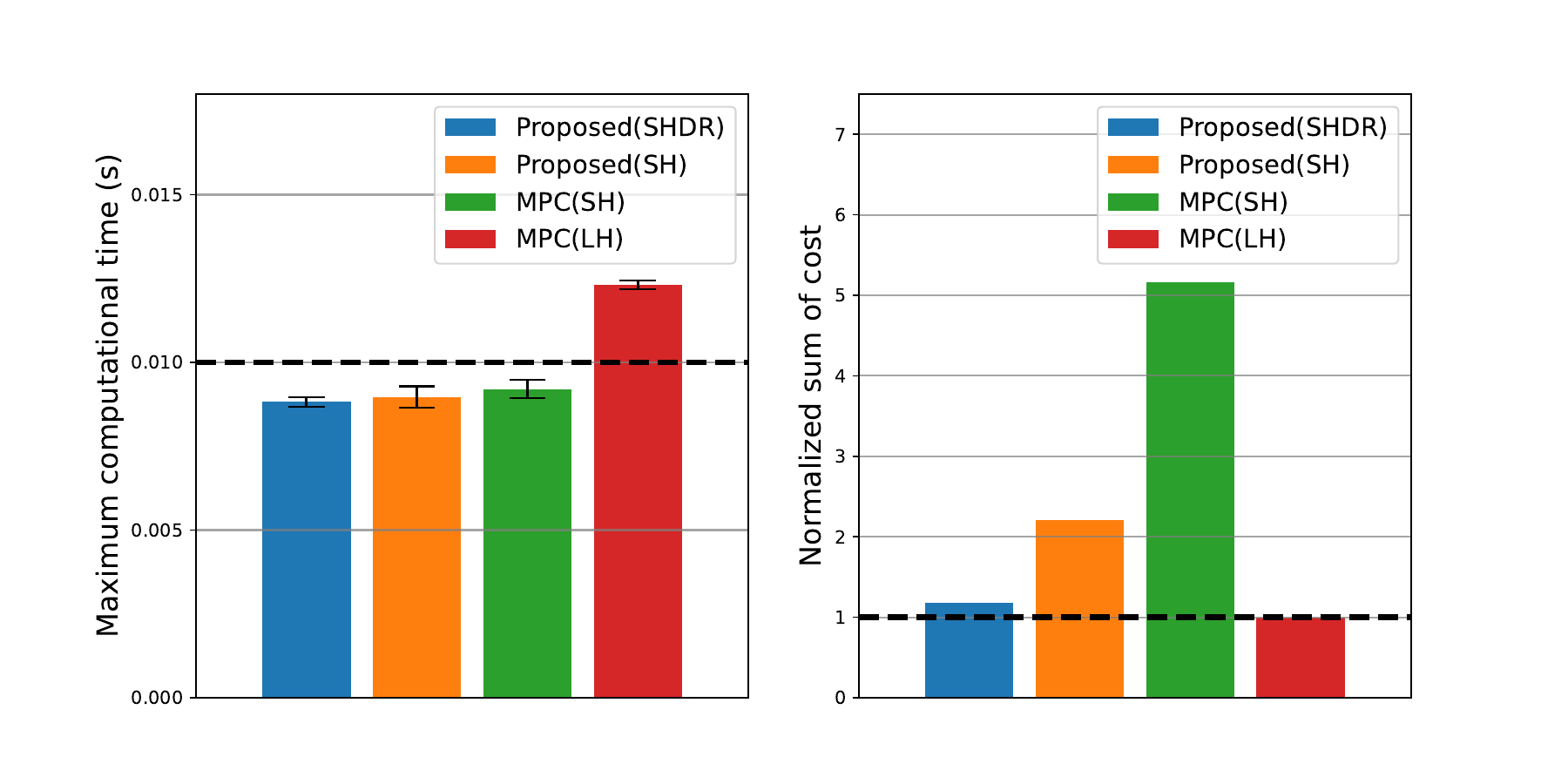}
\caption{The left figure shows the MPC computation time relative to the control cycle. The right figure shows the normalized cumulative cost (lower is better), representing control performance.
MPC(SH) and MPC(LH) represent standard MPC without terminal value learning, with horizons $H=0.4 \, [\si{s}]$ and $H=0.6 \, [\si{s}]$, respectively. MPC(SH) completes computation within the control cycle but results in robot falls and high cumulative costs. MPC(LH) achieves stable performance with low cumulative costs but fails to complete within the control cycle, making it impractical.
MPC(SH) and MPC(SHDR) both use goal-conditioned terminal value learning, with MPC(SHDR) applying domain randomization for uneven terrain. Both have a horizon of $H=0.4 \, [\si{s}]$. From the figures, MPC(SH) reduces cumulative costs through terminal value learning, which were previously high due to falls when not using terminal value. In MPC(SHDR), the robust terminal value learned via domain randomization prevents falls during terrain transitions, which occurred several times in MPC(SH). MPC(SHDR) achieves performance close to MPC(LH) without any falls and completes computations within the control cycle, confirming the effectiveness of the proposed method.}
\label{fig:time_and_cost}
\end{figure}

The goal linear velocity $v_{\text{goal}}$ is set at $1 \ [\si{m/s}]$, the angular velocity $\omega_{\text{goal}}$ at $0 \ [\si{rad/s}]$, and the angles of both the uphill and downhill slopes are set to $15^{\circ}$. The comparison involves the following four methods:
\begin{itemize}
\item Proposed method - Proposed (SHDR)
\item Proposed method (without slope environment randomization) - Proposed (SH)
\item MPC with a short horizon - MPC (SH)
\item MPC with a long horizon - MPC (LH)
\end{itemize}
Here, SH represents Short Horizon, LH represents Long Horizon, and DR represents Domain Randomization. The conditions of MPPI for each method are shown in Table~\ref{comparison_realtime_test}, where the internal model timestep for MPC calculations is set to $2.5 \, [\si{ms}]$.
The results comparing real-time performance and control performance are shown in Fig.~\ref{fig:time_and_cost}.
\begin{table}[hbtp]
\caption{MPPI parameters for each method}
\label{comparison_realtime_test}
\centering
\begin{tabular}{|c|c|c|}
\hline
& Horizon $H \, [\si{s}]$ & Rollouts \\ \hline
Proposed (DR) & 0.4 & 30 \\ \hline
Proposed (without DR) & 0.4 & 30 \\ \hline
MPC (SH) & 0.4 & 30 \\ \hline
MPC (LH) & 0.6 & 35 \\ \hline
\end{tabular}
\end{table}
\\The left figure relates to real-time performance, showing the maximum control computation time relative to the $10 \, [\si{ms}]$ control cycle. The right figure represents the normalized cumulative cost of the task. These results are based on 10 trials each.
From the figure, it's evident that while the MPC with a short horizon can compute in real-time, it incurs a significantly higher cumulative cost, failing to achieve the control objective of traversing the terrain due to falling while ascending the slope. The MPC with a long horizon reduces the cumulative cost but breaks real-time constraints, making it unsuitable for practical applications. 
By setting the control cycle to $13 \, [\si{ms}]$, computations can be completed within the time limit. However, this slows down the feedback control cycle, increasing the likelihood of the robot falling and the task failure rate. For robots with low balance stability, keeping the control cycle as short as possible is especially important.
On the other hand, the proposed method manages to keep the cumulative cost low while maintaining real-time performance, successfully traversing the prepared terrain by dynamically adjusting goal variables based on the slope angle, thus achieving the control objective. Without environment randomization, the robot falls at the transitions of slopes, resulting in higher costs. In fact, falls were observed in 6 out of 10 trials.
However, when domain randomization was applied, the robot did not fall even once in 10 trials. These results highlight the importance of learning terminal values through environment randomization. Hence, it is confirmed that the proposed method can generate a variety of motions while maintaining real-time performance.

\subsection{Evaluation of Robustness}
\begin{figure}[htb]
    \centering 
    \begin{minipage}[b]{0.99\linewidth}
        \includegraphics[width=\linewidth]{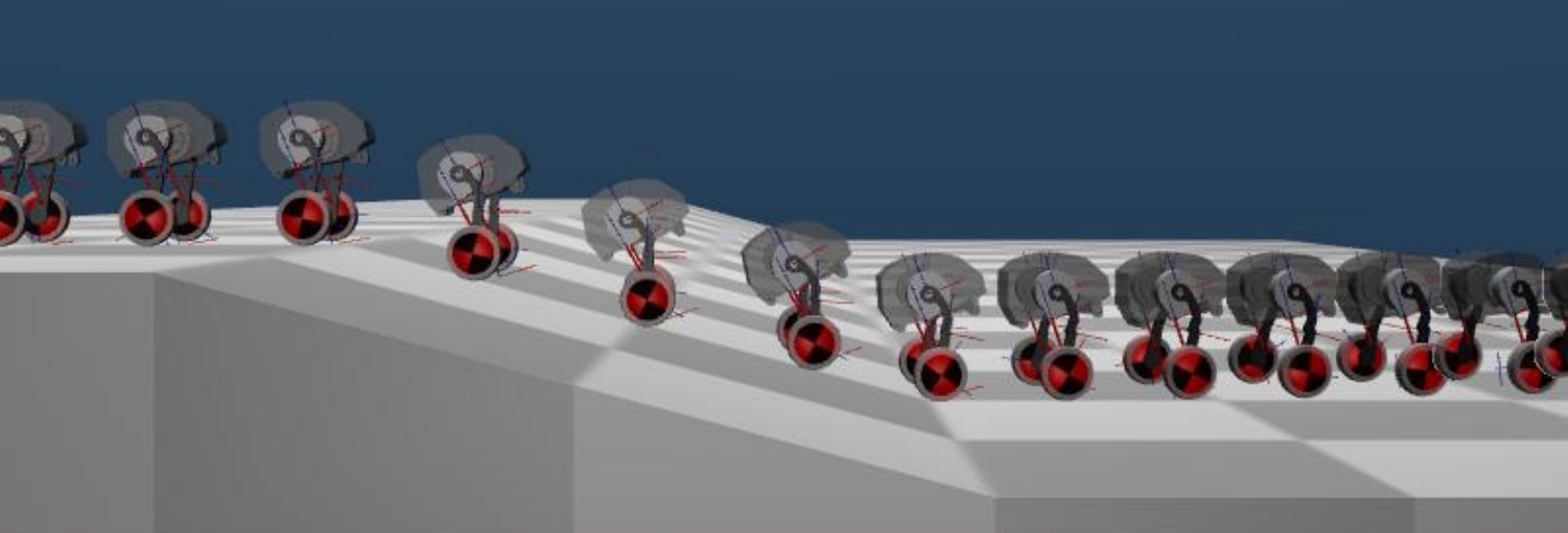}
        \subcaption{Behavior when a force of $20 \, \si{N}$ is applied ($H = 0.4 \, \si{s}$) - Despite being pushed by force, the robot maintains its posture and achieves constant velocity movement.}
        \label{fig:robust_20N}
    \end{minipage} \\
    \begin{minipage}[b]{0.99\linewidth}
        \includegraphics[width=\linewidth]{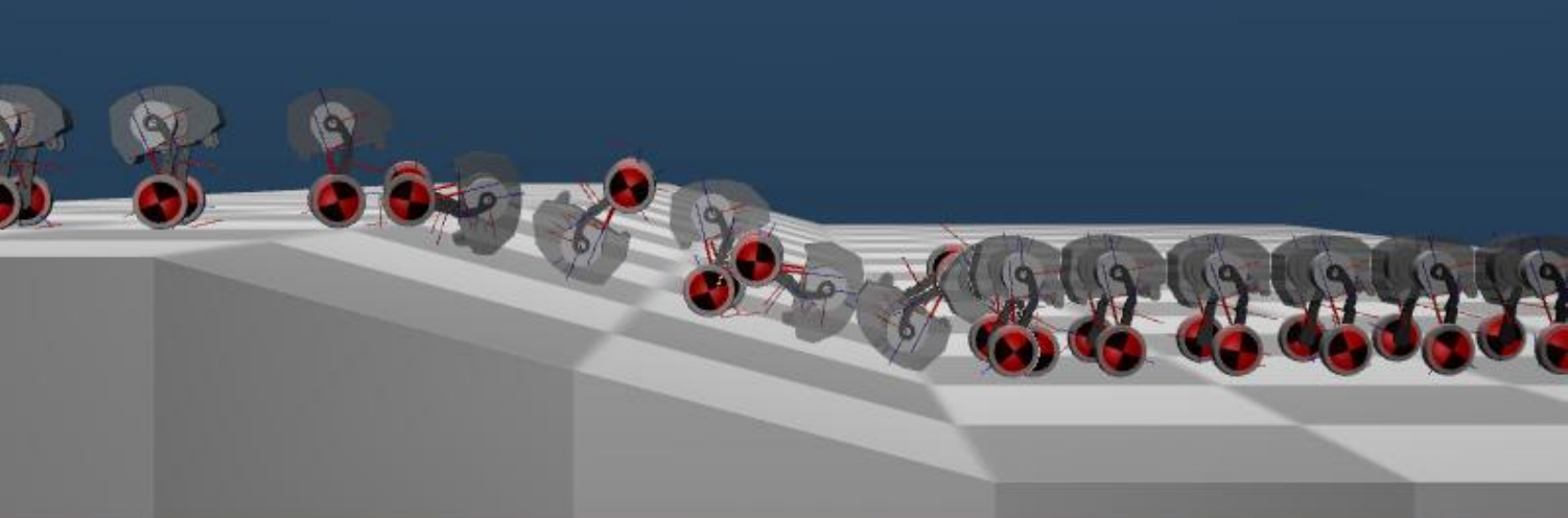}
        \subcaption{Behavior when a force of $40 \, \si{N}$ is applied ($H = 0.4 \, \si{s}$) - Although pushed and rotated by force, the robot subsequently returns to its posture and resumes constant velocity movement.}
        \label{fig:robust_40N}
    \end{minipage} \\
    \begin{minipage}[b]{0.99\linewidth}
        \includegraphics[width=\linewidth]{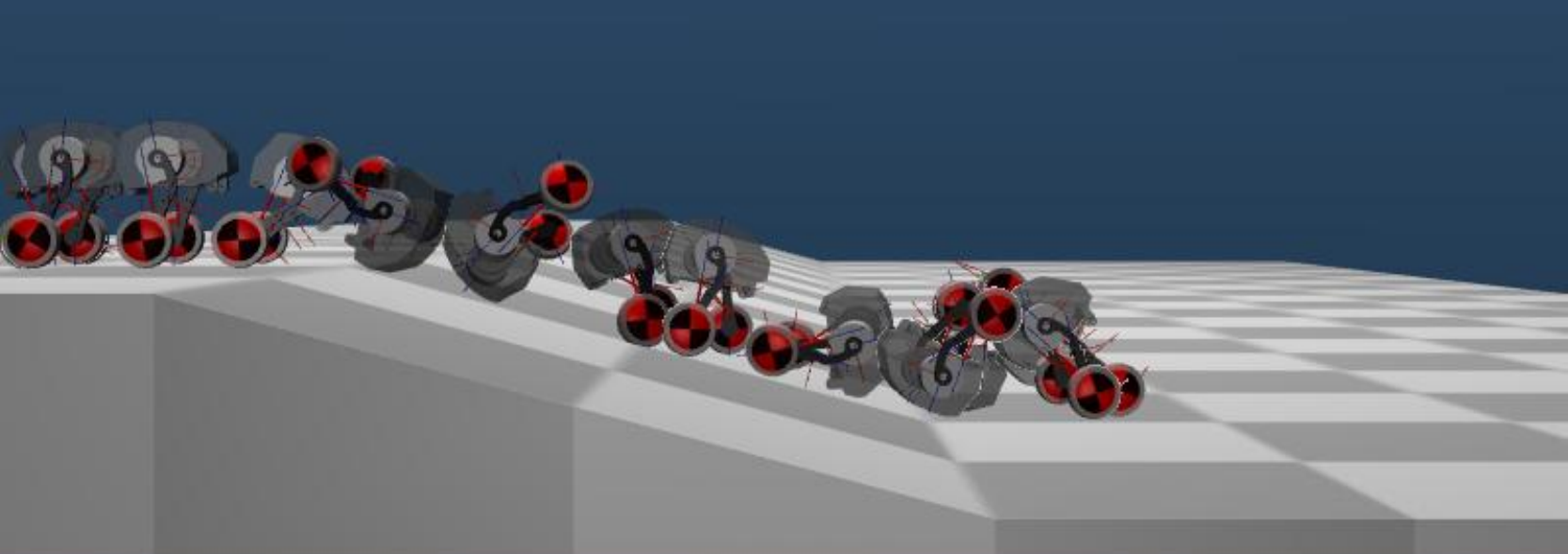}
        \subcaption{Behavior when a force of $40 \, \si{N}$ is applied ($H = 0.24 \, \si{s}$) - With a shorter predictive horizon, the robot fails to recover its posture after rotation and falls.}
    \label{fig:H24_robust_40N}
    \end{minipage}
    \caption{Robot behavior under different forces and predictive horizons.}
    \label{fig:robot_behavior}
\end{figure}

In this section, we verify the robustness of the controller developed through the proposed method by considering scenarios where disturbances are applied during descent on a slope, using the same terrain and conditions described in Section~\ref{subsec:realtime}. The value function learned in Section~\ref{subsec:realtime} is utilized with the proposed method. The behavior of the robot when a force of $20 \, [\si{N}]$ and $40 \, [\si{N}]$ is applied to the base link for 2 seconds during descent is illustrated in Figs.~\ref{fig:robust_20N} and \ref{fig:robust_40N}, respectively.

When a force of $20 \, [\si{N}]$ is applied, the robot descends while maintaining its base link posture despite being pushed. With a force of $40 \, [\si{N}]$, although the robot is pushed and rotates, it regains its posture and continues its motion. By applying the terminal value learned for a horizon of $H = 0.4 \, [\si{s}]$ to an MPC with a horizon of $H = 0.24 \, [\si{s}]$, we employ a method more reliant on the terminal value, similar to reinforcement learning. The result of applying a force of $40 \, [\si{N}]$ in this scenario is shown in Fig.~\ref{fig:H24_robust_40N}. From the figure, it is observed that the robot falls and is unable to return to constant velocity movement after rotation.
\\The recovery motion after rotation observed in Fig.~\ref{fig:robust_40N} was not included during training; it may seem challenging to perform during inference. Unlike reinforcement learning, this behavior is possible because MPC solves an optimization problem at each timestep for a fixed horizon. These results suggest that the proposed method provides a certain level of robustness.

%% file: section/7conclusion.tex
In this study, we addressed the challenge of real-time control for Model Predictive Control (MPC) in high-dimensional, multi-degree-of-freedom systems and complex environments by considering the estimation of terminal values to reduce computational costs. Furthermore, to enable flexible motion generation specific to the task at hand, we proposed a hierarchical MPC framework that learns the terminal values conditioned on the objective, in combination with an upper-level trajectory generator. The proposed method successfully achieved trajectory-following tasks on flat and sloped terrains using a bipedal inverted pendulum robot by shortening the prediction horizon of the MPC and accomplishing tasks in real time. It was also confirmed that terminal values obtained through environmental randomization enhance the robustness of the MPC.

Future challenges include real-world experimentation. Reducing computational costs to enable real-time control can be further validated through experiments with actual robotic systems. We will evaluate how the proposed method, utilizing domain randomization, allows our learning system to achieve successful control of the real robot system.

%% file: section/8appendix_hyperparameter.tex
In this appendix, we describe the various hyperparameters used in the experiments.

We show the hyperparameters of the neural network used for learning the value function in Table \ref{nn_parameter}. For all tasks in this study, the parameters in Table \ref{nn_parameter} were used.
\begin{table}[ht]
\caption{Neural Network Parameters}
\label{nn_parameter}
\centering
\begin{tabular}{|l|l|}
\hline
Parameter Name & Value \\
\hline \hline
Number of hidden layers & 2 \\
Units per layer & 100 \\
Learning rate & 0.001 \\
Replay buffer size & 3000 \\
Minibatch size & 32 \\
Minibatch sampling times & 64 \\
Activation function & ReLU \\
Optimizer & Adam \\
Training frequency $Z$ & $16\ [\si{step}]$ \\
\hline
\end{tabular}
\end{table}

Next, we describe the parameters for MPPI. The parameters for each task are listed in Table \ref{mppi_parameter}. Most parameters are consistent across all tasks, with only the horizon and rollout varying between them.
\begin{table}[hbtp]
\caption{MPPI Parameters}
\label{mppi_parameter}
\centering
\begin{tabular}{|l|l|}
\hline
Parameter Name & Value \\
\hline \hline
Noise $\Sigma$ & $\mathrm{diag}(16,16,1e-5,1e-5,1e-5,1e-5)$ \\
Temperature $\beta$ & 0.1 \\
Discount factor $\gamma$ & 0.99 \\
\hline
\end{tabular}
\end{table}

Lastly, we present the parameters for Kanayama Control in Table \ref{kanayama_params}. $K_x$, $K_y$, and $K_{\psi}$ are the gains for the posture errors in the $x$ direction, $y$ direction, and yaw direction, respectively.
\begin{table}[hbtp]
\caption{Kanayama Control Parameters}
\label{kanayama_params}
\centering
\begin{tabular}{|l|l|}
\hline
Parameter Name & Value \\
\hline \hline
$K_x$ & 3.0 \\
$K_y$ & 3.0 \\
$K_{\psi}$ & 2.0 \\
\hline
\end{tabular}
\end{table}